\documentclass{article}

    \PassOptionsToPackage{numbers, compress}{natbib}
    \usepackage[preprint]{neurips_2025}

\usepackage[utf8]{inputenc} % allow utf-8 input
\usepackage[T1]{fontenc}    % use 8-bit T1 fonts
\usepackage{hyperref}       % hyperlinks
\usepackage{url}            % simple URL typesetting
\usepackage{booktabs}       % professional-quality tables
\usepackage{amsfonts}       % blackboard math symbols
\usepackage{nicefrac}       % compact symbols for 1/2, etc.
\usepackage{microtype}      % microtypography
\usepackage{xcolor}         % colors

\usepackage{amsmath,amsfonts,bm}

\def\eqref#1{equation~\ref{#1}}
\def\1{\bm{1}}

\def\va{{\bm{a}}}

\def\vh{{\bm{h}}}

\def\vo{{\bm{o}}}

\def\vs{{\bm{s}}}

\def\vx{{\bm{x}}}

\DeclareMathAlphabet{\mathsfit}{\encodingdefault}{\sfdefault}{m}{sl}
\SetMathAlphabet{\mathsfit}{bold}{\encodingdefault}{\sfdefault}{bx}{n}

\usepackage{hyperref}
\usepackage{url}
\usepackage{amsmath,graphicx,amssymb,mathtools,amsthm,float,mathtools}
\usepackage{subcaption}
\usepackage{multirow}
\usepackage{booktabs}
\usepackage{wrapfig}
\usepackage{makecell}
\usepackage{dsfont}
\usepackage{algorithm}
\usepackage{algpseudocode}

\theoremstyle{plain}
\newtheorem{theorem}{Theorem}[section]

\newtheorem{lemma}[theorem]{Lemma}

\theoremstyle{remark}

\definecolor{mydarkblue}{rgb}{0,0.08,0.45}
\definecolor{mydarkgreen}{RGB}{0, 139, 69} 
\hypersetup{
	colorlinks=true,
	urlcolor=magenta,
	citecolor=mydarkblue,
}

\title{ManiBox: Enhancing Embodied Spatial Generalization via Scalable Simulation Data Generations}

\author{%
  Hengkai Tan$^{1}$\thanks{Equal contribution},~
  Xuezhou Xu$^{2*}$,~ 
  Chengyang Ying$^{1*}$,~
  Xinyi Mao$^{1}$,~
  Zeyuan Wang$^{1}$,\\
  \textbf{Songming Liu}$^{1}$,~
  \textbf{Xingxing Zhang}$^{1}$,~
  \textbf{Zhizhong Su}$^{3}$,~
  \textbf{Hang Su}$^{1}$, ~
  \textbf{Jun Zhu}$^{1}$
   \\
  $^{1}$Tsinghua University
  $^{2}$National University of Singapore
  $^{3}$Horizon Robotics \\
  \texttt{thj23@mails.tsinghua.edu.cn,xu.xuezhou@u.nus.edu,ycy21@mails.tsinghua.edu.cn} \\
}

\begin{document}

\maketitle

\begin{abstract}
Embodied agents require robust spatial intelligence to execute precise real-world manipulations. However, this remains a significant challenge, as current methods often struggle to accurately position objects in space.
Collecting extensive data can help address this issue by enhancing the agent's spatial understanding. Nonetheless, obtaining such data with real robots is prohibitively expensive, and relying on simulation data frequently leads to visual generalization gaps during real-world deployment. 
% We first show that the challenge of spatial generalization stems primarily from the extensive data needed for spatial understanding. 
% However, collecting such data with real robots is prohibitively expensive, and relying on simulation data often leads to visual generalization gaps during real-world deployment.
To tackle these challenges, we propose \textbf{ManiBox}, a novel bounding-box-guided framework. By decoupling perception from policy generalization, ManiBox effectively reduces the Sim2Real gap, leverages Internet-scale data, and scales our policy data collection in simulation.
% To address these challenges, we propose \textbf{ManiBox}, a novel bounding-box-guided framework, decoupling observation perception and policy generalization, 
% which helps reduce the Sim2Real gap,  and scale up our collected simulation data.
% which helps reduce the Sim2Real gap, leverage Internet-scale data, and scale our collected policy data in the simulator. 
% In detail, we propose \textbf{ManiBox}, a novel bounding-box-guided method built on a simulation-based teacher-student framework.
Specifically, within ManiBox, the RL teacher policy efficiently generates scalable simulation data. The student policy is distilled from this data and takes bounding boxes as input, which is proven sufficient for determining objects' spatial positions, thus enabling zero-shot transfer to real robots.
% Specifically, within ManiBox, the RL teacher policy efficiently generates scalable simulation data, and the student policy inputted with bounding boxes, which are proven sufficient to determine the objects' spatial positions, is distilled for zero-shot transferring to real robots.
Comprehensive evaluations in both simulated and real-world environments demonstrate that ManiBox exhibits strong spatial generalization and adaptability across various manipulation tasks and settings.
Furthermore, our empirical study provides preliminary verification of spatial scaling laws, i.e., the amount of data required for spatial generalization scales with spatial volume following a power-law relationship.
At a given spatial volume level, the success rate of manipulation tasks follows Michaelis-Menten kinetics with respect to data volume, exhibiting a saturation effect as data increases.\footnote{Our videos and code are available in the \href{https://thkkk.github.io/manibox}{project page}}
\end{abstract}

\begin{figure*}[t]
\begin{center}
%\framebox[4.0in]{$\; $}
% \fbox{\rule[-.5cm]{0cm}{4cm} \rule[-.5cm]{12cm}{0cm}}
\includegraphics[width=0.9\linewidth]{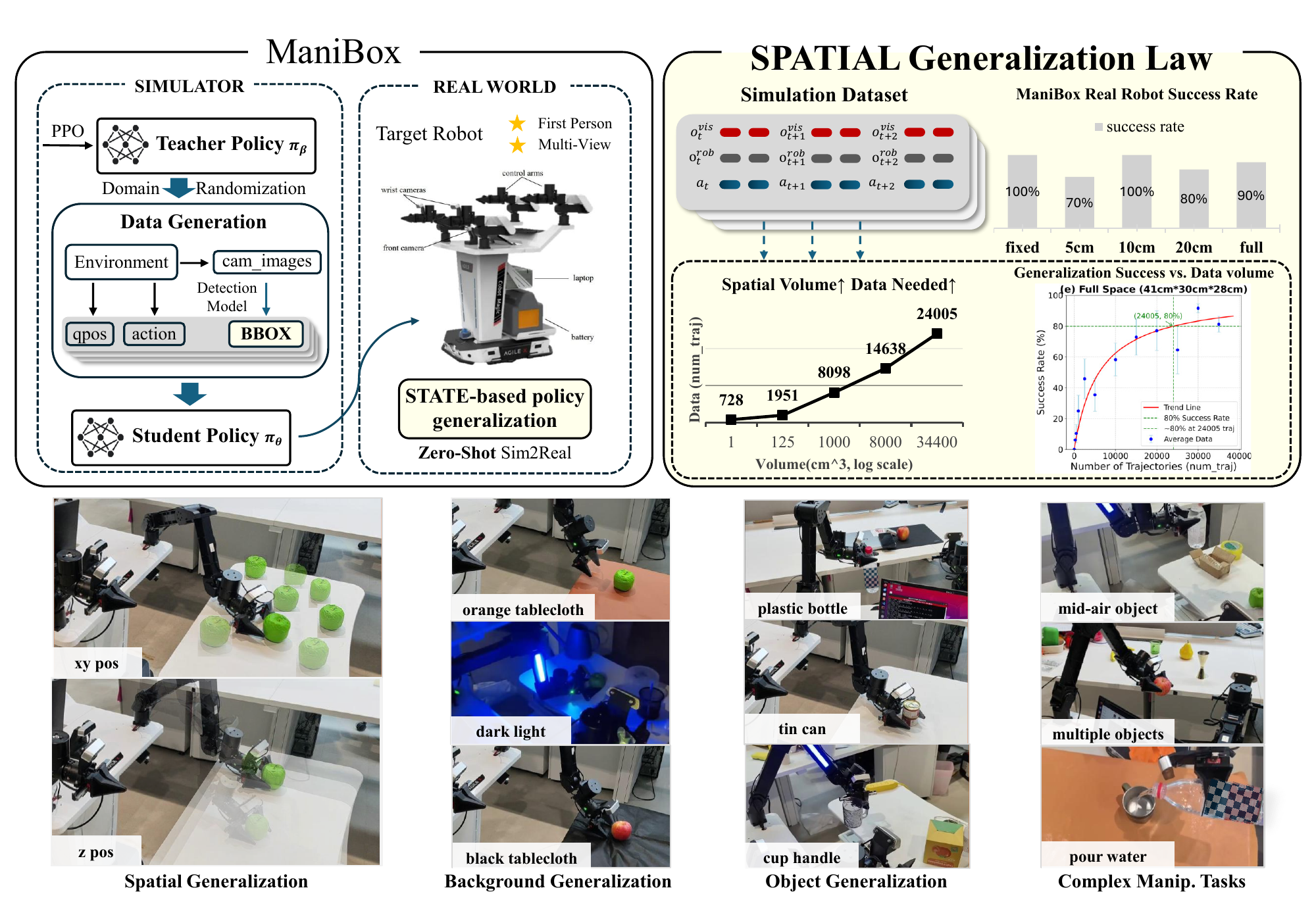}
\end{center}
\vspace{-0.5cm}
% \caption{header image.\hangx{highlight the contribution}}
\caption{\textbf{Overview.}
We introduce ManiBox, a bbox-guided manipulation method using a teacher-student framework to enhance embodied spatial generalization. We reveal that spatial volume generalization scales positively with data volume in a power law, with manipulation success following Michaelis-Menten kinetics relative to data volume for specified spatial volumes. Extensive real-world tests show ManiBox's robust adaptability to varied spatial positions, objects, and backgrounds. 
} 
\vspace{-0.75cm}
\label{fig:header}
\end{figure*}

\section{Introduction}

Spatially accurate robotic manipulation in dynamic environments is crucial for advancing modern robotics~\citep{fu2024mobile}. 
Enabling precise manipulation in unstructured settings extends robots' utility beyond static environments and enhances their real-world effectiveness, particularly in industrial and domestic applications.
% Equipping robots with precise manipulation capabilities in unstructured settings not only extends their utility beyond traditional static scenarios but also enhances their real-world effectiveness, especially in industrial and domestic settings.
A key requirement for such systems is the development of \textbf{spatial generalization}—\emph{the ability of a manipulation agent to complete tasks regardless of the target object's position within a defined spatial volume}. 
This generalization is crucial for varying real-world spatial conditions, where the objects are rarely fixed in place.
% This generalization to perform successfully across various spatial conditions is essential, as objects in real-world settings are rarely fixed in place, and effective manipulation demands adaptability to a broad range of spatial challenges.

Recent work has made notable progress in embodied manipulation in different tasks or environments, such as large embodied foundation models driven by end-to-end training~\citep{ahn2022can,shridhar2022cliport,brohan2022rt,shridhar2023perceiver,zitkovich2023rt,kim2024openvla}. 
However, these models still lack sufficient spatial awareness and reasoning capabilities, posing substantial challenges for spatial generalization in real-world applications~\citep{cheng2024spatialrgptgroundedspatialreasoning}. 
Our analysis (Lemma~\ref{lemma1}) suggests that achieving spatial generalization over larger volumes requires substantially more data, a relationship we also empirically verify as a power law (Figure~\ref{fig:data_vs_volume}). 
For example, training a grasping policy at a fixed point may only require 50 to 600 trajectories. However, when the grasping policy need to generalize within a spatial volume of $34,400 \mathrm{cm}^3$ ($41 \mathrm{cm} \times 30 \mathrm{cm} \times 28 \mathrm{cm}$, roughly the maximum reach of our robotic arm), the required data increases $34400^{0.35}=38$ times, demanding tens of thousands of trajectories.

Unfortunately, existing robotic action trajectories for embodied foundation models is significantly more scarce on the internet compared to image datasets.
And collecting such large data on real robots to achieve spatial generalization is prohibitively expensive. For example, RT-1 spends 17 months accumulating 130K episodes~\citep{brohan2022rt}, a level of resource commitment that is unfeasible for most researchers. 
Although simulation data can help bridge this gap, significant high-dimensional visual discrepancies between simulated and real environments, known as the Sim2Real gap, also pose additional challenges. 
While prior efforts have aimed to enhance performance using various visual generalization techniques, each method has limitations. 
Some require unobstructed third-person perspectives~\citep{yuan2024learning}, while others demand substantial computational resources~\citep{shridhar2023perceiver, ahn2022can}.
These constraints hinder real-time processing and practical deployment on resource-limited robotic platforms.
% Some demand unobstructed third-person perspectives~\citep{yuan2024learning}, some consume substantial computational resources~\citep{shridhar2023perceiver, ahn2022can}.
% These requirements can impede real-time processing and practical deployment on resource-limited robotic platforms.

%ycy:我们解耦了视觉感知和策略泛化，有两个主要的好处，一个是视觉感知可以借助已有的图像模型（yolo-world等），从而避免采集大量的数据实现视觉泛化，另一个是对于视觉感知不是很准的地方，我们通过策略泛化，提高了鲁棒性
% To address this gap, we propose decoupling the observation perception and policy generalization, which is beneficial for reducing the Sim2Real gap and improving the spatial generalization. 
To bridge this gap, we propose decoupling observation perception from policy generalization, which helps reduce the Sim2Real gap and enhance spatial generalization.
First, instead of training image-based policies, we choose visual intermediate representations like bounding-box as the policy inputs. This choice can utilize internet-scale datasets to improve visual representation learning and significantly reduce the Sim2Real gap compared with the original images.
Although visual intermediate representations may lose some information, our lemma~\ref{lemma2} demonstrates the completeness of bounding boxes captured from multiple cameras; that is, they can effectively encapsulate the 3D structure of convex objects. 
Moreover, taking visual intermediate representations as the input is beneficial for \textbf{scaling up the collected action data in simulators} and improving the policy generalization by reducing the Sim2Real gap, which can also facilitate spatial generalization, even when the visual representations are biased.
Based on these observations, we propose a novel bounding-box-guided methodology \textbf{ManiBox} (Figure~\ref{fig:header}) to enhance spatial generalization and adaptability.
In detail, ManiBox first trains a teacher policy with privileged information in the simulator with reinforcement learning (RL). This policy then generates scalable robot trajectories, replacing traditional visual inputs with bounding box coordinates to facilitate spatial generalization of manipulation policies. 
Subsequently, the student policy trained on these simulation data achieves robust zero-shot transfer to real-world tasks, leveraging bounding boxes identified by advanced open-vocabulary detection models, such as YOLO-World \citep{cheng2024yolo}.

% While perception models are essential for object and environmental recognition, the true effectiveness of a robot in dynamic settings hinges on the robustness of its policy execution. A well-generalized policy empowers the robot to adjust its actions in real time, effectively compensating for visual inaccuracies and adapting to changes in object positions or environmental dynamics. Such adaptability is crucial for achieving task success and maintaining system stability, especially in the face of imperfect or noisy perception data.
% To further enhance the robot's ability to manage unpredictable scenarios (e.g., Sec.~\ref{sec:random_mask_inference}) and ensure reliable real-world performance, we shift our focus to \textbf{state-based policy generalization} instead of relying solely on vision-based methods. Our Lemma~\ref{lemma2} demonstrates that bounding boxes, when captured from multiple cameras, effectively encapsulate the 3D structure of convex objects, providing an optimal low-dimensional state for policy generalization. Building on this foundation, we introduce a novel bounding-box-guided methodology \textbf{ManiBox} to enhance spatial generalization and adaptability.

Extensive experiments in both simulators and real robots demonstrate a direct correlation between data volume and spatial generalization performance: \emph{more data consistently leads to higher success rates, and generalizing to larger spatial volumes requires even more data}. 
The success rate and the data volume show the \textbf{Michaelis-Menten kinetic curve}. 
And the data volume required for spatial generalization is related to the spatial volume in a \textbf{power-law relationship}.
For example, to reliably grasp objects within a $34400 cm^3$ workspace, tens of thousands of trajectories from various positions are required. 
ManiBox achieves nearly perfect success rates in real-world grasping tasks by leveraging large-scale simulation data generated across diverse spatial volumes. 
Furthermore, ManiBox is flexible for handling different manipulation tasks and exhibits robust generalization in new scenarios, including unfamiliar object types, complex surfaces, changing lighting conditions, and environments with distractions.

Our contributions include: 
(1) %We propose to decouple visual representation and policy generalization for reducing the Sim2Real gap as well as scaling up the simulation data, resulting in our \textbf{ManiBox} for spatial generalization in real-world manipulation; 
We propose \textbf{ManiBox} , a novel framework that decouples visual representation from policy generalization to reduce the Sim2Real gap and scale up simulation data;
(2) %ManiBox significantly enhances adaptability to diverse spatial positions, object types, and backgrounds, as demonstrated by extensive experiments in both simulated and real-world environments; 
ManiBox significantly enhances adaptability across diverse spatial positions, object types, and backgrounds, as demonstrated by extensive evaluations in both simulated and real-world environments. 
(3) %Through scalable simulation data generation, we provide novel insights into the relationship between data volume and spatial generalization, establishing a framework for improving generalization across a variety of embodied systems.
We take an initial step in unveiling the impact of data volume on spatial generalization through scalable data generation, laying the foundation for improving generalization in embodied systems.

\section{Related Work}
% \paragraph{Generalization for Robotic Manipulation.}
\textbf{Generalization for Robotic Manipulation.}
Generalization is fundamental and significant for robotic manipulation, especially in dynamic environments with varying visual inputs and changing conditions. Existing vision-based policy generalization methods leverage both simulated and real-world data. By simulating diverse visual scenarios, robots can better generalize to new environments and tasks. 
For instance, multi-task and multi-object manipulation studies have extended behavior cloning to handle different visual inputs, improving generalization to unseen objects and scenes~\citep{vuong2023grasp,fang2023anygrasp}. Additionally, language-conditioned policies~\citep{brohan2022rt,zitkovich2023rt,shridhar2023perceiver} and imitation learning from human videos~\citep{chen2021learning,nair2023r3m} enable robots to manipulate objects using both visual and language cues.
% However, scaling real-world demonstrations to improve generalization remains challenging due to resource constraints. 
% Simulation-based learning mitigates this by offering large, diverse visual data. 
% Domain randomization techniques~\citep{tobin2017domain,peng2018sim} have proven effective in generating varied visual conditions, enabling models to generalize more robustly across real-world tasks. 
While methods like bounding box annotations assist in specific tasks, such as pose estimation~\citep{huang2023earl}, understanding the relationship between dataset scale and generalization remains an area that requires further exploration. 
% As simulation data grows, more research is needed to evaluate how data volume impacts policy generalization, particularly in dynamic environments.

% \paragraph{Embodied Learning from Simulation.}
\textbf{Embodied Learning from Simulation.}
Simulators provide an efficient and scalable way to expand embodiment datasets, offering a cost-effective means for parallel data collection and faster iteration cycles. There are three main approaches: 1) training RL agents through interaction within the simulator~\citep{huang2023earl,radosavovic2024real,yuan2024learning}, 2) mimicking expert demonstrations generated in simulation with imitation learning~\citep{xie2020deep,tan2024fourier,mu2024robotwin}, and 3) employing teacher-student frameworks, i.e., training a teacher policy with privileged information to guide a student policy~\citep{eth2020,geng2023partmanip,zhuang2023robot}. 
% These methods reduce the reliance on expensive real-world data collection.
One of the major challenges of utilizing simulation data is the Sim2Real gap.
Various techniques have been developed to bridge it, including realistic simulators~\citep{todorov2012mujoco,james2020rlbench,makoviychuk2021isaac}, domain randomization~\citep{tobin2017domain,peng2018sim,zhuang2023robot}, and vision-based policy generalization~\citep{geng2023partmanip,tan2024fourier,ying2024peac,yuan2024learning}. These approaches enable robots to generalize across diverse visual and physical conditions, which are critical in real-world environments. 
% For example, domain randomization allows policies trained in varied simulated conditions to transfer effectively to real-world scenarios~\citep{tobin2017domain,peng2018sim}. 
While real-world datasets such as Open X-Embodiment~\citep{padalkar2023open} provide 1M demonstrations, they are much smaller than datasets in other domains like language and images~\citep{lehmann2015dbpedia,muhleisen2012web,weyand2020google,wu2019tencent}. 
Thus, simulation-generated datasets play a vital role in compensating for the limitations of real-world data collection.

\begin{figure*}[t]
\begin{center}
%\framebox[4.0in]{$\; $}
% \fbox{\rule[-.5cm]{0cm}{4cm} \rule[-.5cm]{12cm}{0cm}}
\includegraphics[width=\linewidth]{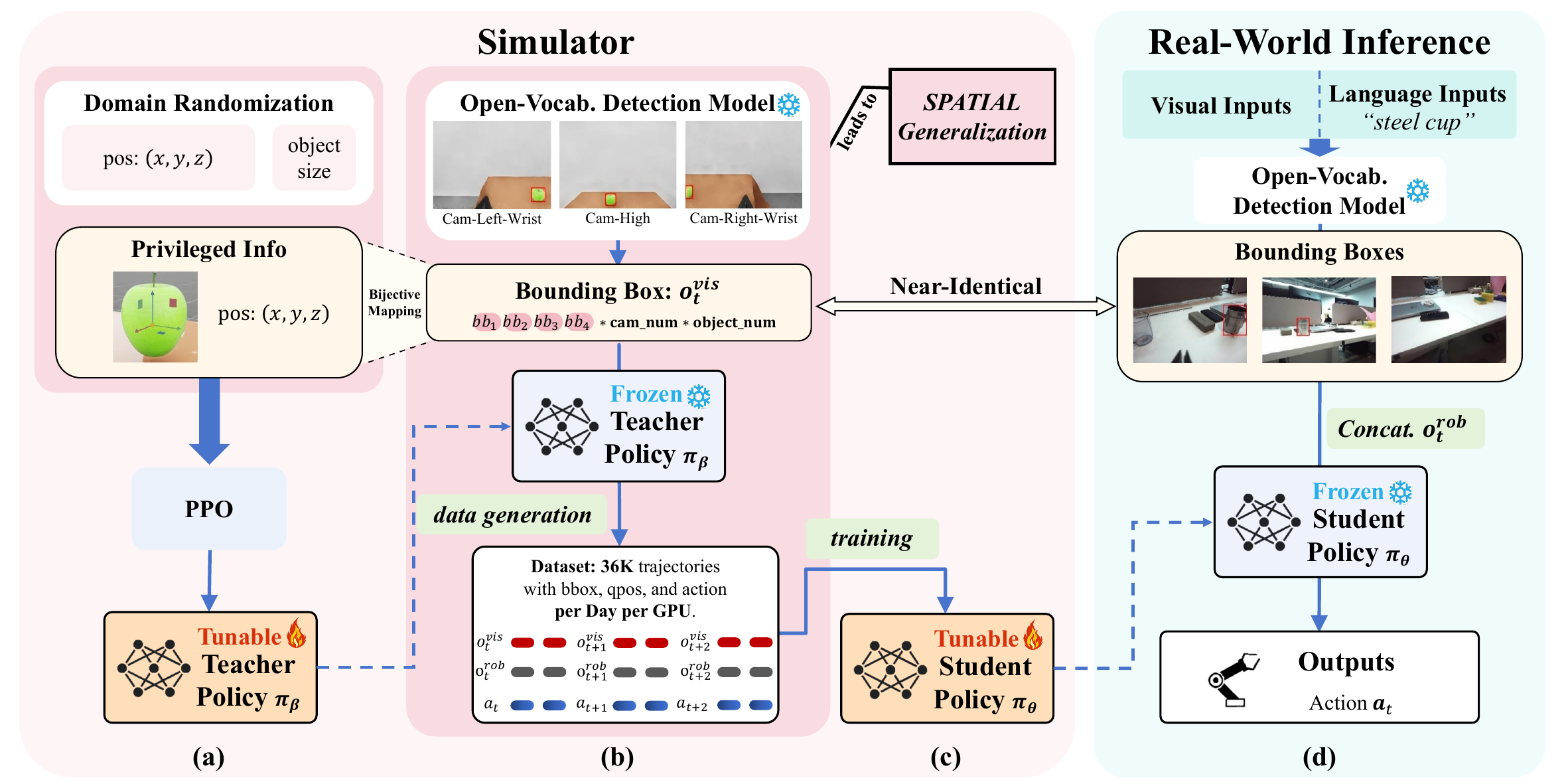}
\end{center}
% \caption{Method overview. We use a teacher-student approach, training the teacher policy in the simulator with domain randomization, privileged information, and PPO (left column). The teacher policy then generates many robot trajectories based on bounding boxes (center column), which are used to train the student policy for real-world deployment (right column). The bounding boxes are created with YOLO-World to capture key visual information.}
% \caption{The illustration of our teacher-student framework for grasping policy training and deployment. Initially, the teacher policy is trained within a simulated environment using domain randomization and privileged information, employing PPO as depicted in the left column. This policy then generates a dataset comprising numerous robot trajectories informed by object-bounding boxes created with YOLO-World (shown in the left column). Subsequently, the generated data is utilized to train a student policy, which is designed for deployment in real-world scenarios (illustrated in the right column). The student policy integrates visual as well as linguistic inputs, processes them, and outputs the robot's actions. \thk{to be modified...}}
\vspace{-0.2cm}
\caption{\textbf{ManiBox illustration.} 
% \textbf{The illustration of ManiBox for grasping policy training and deployment.}
(a) Utilizing PPO and domain randomization, we train a 
sophisticated teacher policy in the simulator that utilizes privileged object information to determine actions.
% We use domain randomization, provide privileged information, and employ PPO to train a teacher policy with good spatial generalization in the simulator.
(b) The teacher policy generates scalable trajectory data, utilizing bounding box coordinates instead of traditional high-dimensional visual inputs or privileged information.
% This policy then generates a dataset comprising large-scale robot trajectories containing object-bounding boxes created with an open vocabulary detection model, as well as the joint position and action of the robot.
(c) A state-based student policy, generalizable and capable of zero-shot transfer, is trained on this extensive simulation dataset, greatly improving spatial generalization.
% The generated data is utilized to train a student policy specifically designed for deployment in real-world scenarios.
(d) Guided by bounding boxes, the student policy precisely executes actions for real robots, achieving improved generalization capabilities.
% The student policy integrates visual as well as linguistic inputs, processes them, and outputs the robot's actions.
%\hangx{Important! try to explain why and how we improve the spatial generalization using the pipeline. Try to make it explictly using the figure and caption.}\xx{already revised.}
}
\vspace{-0.5cm}
\label{fig:method}
\end{figure*}

\section{Methodology}

In this section, we first formulate the problem and then introduce our ManiBox, including teacher policy training, simulation data generation, and student policy distillation. 
% Our methods aim to increase the generalization of the control policy, including the generalization of position, scenarios, and objects. ,,

% \mxy{What do you mean by 'decouple data collection and real-world policy training'? Somehow confused. Maybe 'decouple data collection and real-world environment' or so?}

\subsection{Problem Formulation}
\label{sec_problem}

For robotic manipulation within dynamic environments, achieving spatial generalization (i.e., the ability to effectively perform tasks across varied spatial configurations) is essential for usability and overall efficacy. 
We formulate it as a partially observable Markov decision process (POMDP)~\citep{cassandra1994acting} as the robot cannot access to complete state information, especially in real-world settings.
Formally, we consider the POMDP $\mathcal{M} = (\mathcal{S}, \mathcal{A}, \mathcal{O}, \mathcal{T}, \mathcal{R}, \gamma)$, where $\mathcal{S}$ represents the state space, including all configurations of the robot and its environment;
% all possible robot and environment configurations;
$\mathcal{A}$ denotes the action space, containing all possible robot actions; $\mathcal{O}$ is the observation space, which reflects the limited state information accessible to the robot; $\mathcal{T}(\vs_{t+1}|\vs_t, \va_t)$ and $\mathcal{R}(\vs_t, \va_t)$ are the state transition dynamics; and the reward function, respectively; and $\gamma$ is the discount factor.
%  modulating the significance attributed to immediate versus future rewards. 
% which controls the importance of future rewards.

In real-world scenarios, the robot receives partial observations $\vo_t \in \mathcal{O}$, which provides incomplete and potentially noisy information about the true state $\vs_t$. 
% The observation can be represented as  $\vo_t \sim \mathcal{O}(\vs_t)$.
The robot's objective is to find a policy $\pi$ that maps observations (or histories of observations) to actions $\va_t \sim \pi(\va_t | \tau_t)$, where $\tau_t = (\vo_0, \va_0, \vo_1, \ldots, \vo_t)$ is the history. The policy requires maximizing the expected cumulative reward:
\begin{equation}
   J(\pi) = \mathbb{E}_{\tau\sim\pi} \left[ \sum_{t=0}^{\infty} \gamma^t \mathcal{R}(\vs_t, \va_t) \right].
\end{equation}
% where the expectation is over the trajectories induced by the policy $\pi$. 
% However, solving POMDPs optimally is computationally intractable due to the need to maintain a belief over the state space. To make the problem tractable, we employ a teacher-student framework.
% However, optimally solving Partially Observable Markov Decision Processes (POMDPs) is known to be computationally intractable, as it requires maintaining a belief distribution over the entire state space. 
% \junz{very brief on previous efforts? to contract with our choice...}\hang{
However, solving POMDPs for manipulation tasks is computationally expensive, as it requires the belief distribution over all possible states, including the precise position, orientation, and object dynamics. This complexity is compounded by the need to handle occlusions, sensor noise, and dynamic changes in the environment, making it challenging for robots to operate efficiently in real-world settings where the state is constantly evolving and uncertain \citep{kaelbling1998planning,kochenderfer2015decision}.%}

% To mitigate this complexity\junz{too general? any specific challenge over existing ones? otherwise, it looks straightforward/arbitrary to choose the T-S framework...}, 
To reduce this complexity and improve the real-world spatial generalization, we develop \textbf{ManiBox}, a novel bounding-box-guided manipulation method by adopting the teacher-student framework~\citep{eth2020,geng2023partmanip}.
First, ManiBox enables efficient, scalable data generation, as the teacher policy in simulation can produce a large and diverse set of trajectories. 
Moreover, with low-dimensional bounding box states, Manibox effectively reduces the Sim2Real gap and accelerates the training of the student policy. 

However, due to the complexity of real-world manipulation, there are several notable challenges in designing Manibox. First, the teacher policy must efficiently explore high-dimensional state spaces in simulation, which are computationally intensive. The student policy needs to handle incomplete observations and adjust to real-world variations, such as sensor noise and dynamic environmental conditions not present in simulations. Below, we will introduce core concepts of Manibox for overcoming these challenges and show that Manibox is flexible to handle different manipulation tasks in our experiments.
The pseudo algorithm of ManiBox can be seen in Alg.~\ref{alg:manibox}. 

\subsection{Teacher Policy Training}
\label{sec_teacher}
% \hangx{why we need teacher policy? what is the purpose for teacher policy?what are the challenges for training a teacher policy?}

% The goal of the teacher policy training is to explore a teacher policy $\pi_\beta: \mathcal{S} \rightarrow \mathcal{A}$, maximizing the cumulative reward 
% \begin{equation}
%     \mathbb{E}_{\tau \sim \pi_\beta} [R(\tau)]=\mathbb{E}_{\tau \sim \pi_\beta} \left[\sum_{i=0}^{\infty} \gamma^{i} r(s_i, a_i)\right]
% \end{equation}
% in simulator. By properly designing the reward function, the obtained teacher policy has the ability to grasp objects.

% For a scalable approach to production data, we first train the teacher policy $\pi_\beta: \mathcal{S} \rightarrow \mathcal{A}$ in the simulation to maximize the reward by utilizing all state information, including the privileged information $s_{\text{pri}}$.
% As a result, expert policies can automatically generate a substantial amount of data even without the need for human assistance, such as teleoperation.
% 如果是“substantial”+“a single GPU”的话，看起来更像是资源少与数据多的、数量上的对比mxy
To automatically and efficiently generate a large amount of diverse data, we first train an expert \textbf{teacher policy} $\pi_\beta: \mathcal{S} \rightarrow \mathcal{A}$ in the simulation where it can access the full state information $\vs_t$, including privileged information such as precise object positions. We optimize $\pi_{\beta}$ with Proximal Policy Optimization (PPO)~\citep{schulman2017proximal} by maximizing the expected reward $J(\pi_{\beta})$,
% \begin{equation}
%     J(\pi_{\beta}) = \mathbb{E}_{\tau\sim\pi_{\beta}} \left[ \sum_{t=0}^{\infty} \gamma^t \mathcal{R}(\vs_t, \va_t) \right],
% \end{equation}
%The teacher policy generates a diverse set of trajectories across a wide range of spatial configurations, learning optimal actions in simulation that will later guide the student policy.
% where the reward function is defined to guide the policy in grasping the object and moving it to a specified location. 
where the reward function is designed to handle the manipulation task. 
Especially, there are several advantages of the choice of lower-dimensional states in teacher policy: benefiting efficient learning, reducing exploration space through privileged information, and supporting larger parallel environments
To leverage these benefits, we carefully design the simulation environment to reduce the exploration space and make it easier for the policy $\pi_\beta$ to find the point with the highest return below
(More details are in Appendix~\ref{app_simulation}):

% Our teacher can generate a diverse set of optimal trajectories across a wide range of spatial configurations for further training the real-world student policies. 
% Below, we will introduce the core concepts of Manibox and show that Manibox is flexible to handle different manipulation tasks in our experiments. 
% These trajectories serve as training data for the student policy, guiding it to learn effective manipulation policies.

% \revise{The advantages of using a teacher policy instead of direct visual RL training can be summarized in three key aspects. 
% First, the teacher policy relies solely on robot and object information as input, significantly reducing the dimensionality of the state space. 
% Second, the incorporation of privileged information and tailored reward functions for critical steps offers more focused guidance during training, thereby narrowing the exploration space. Lastly, visual RL suffers from limited parallelizability, which results in lower overall training efficiency.
% } 

\textbf{Environment Setup.} To ensure efficient policy training and data generation, we utilize Isaac Lab~\citep{mittal2023orbit}, which supports large-scale parallelism environments. 
The scene includes a platform, a robot, and objects arranged simply to facilitate future domain randomization. 
% The robot's goal is to complete the corresponding manipulation task. 
We apply the dual-arm Mobile ALOHA~\citep{fu2024mobile} robot, equipped with three first-person-view RGB cameras, into the simulator and ensure its dynamics are consistent with the real world.
% To support future extensions to more complex tasks, such as obstacle avoidance during grasping, we represent the robot's behavior with a 30Hz trajectory. 
% To support real-world inference, we represent the robot's behavior with a 30Hz trajectory. 

\textbf{State.} As mentioned above, our teacher policy take the full state information $\vs_t$ as the input, including privileged information, which can accelerate RL training and benefit learning generalized policies.
% For training efficiency, our teacher policy is state-based and can access the full state information $\vs_t$, including privileged information like the object's position. %$\vs_{\text{pri}}$. 
% This privileged information can accelerate RL training and enable the agent to learn more efficient and generalized policies. 
% It is particularly useful in teacher-student frameworks, where the teacher policy utilizes privileged information to guide the student's learning process.

% Besides, the rest of the state includes proprioceptive data $\vs^{\text{rob}}$, such as joint position and joint velocity. As for the action, we choose position control which is the expected joint position for the next timestep.

% \paragraph{Reward.} To ensure the policy achieves the desired grasping behavior on our robot, we design a reward function that reflects the task's objectives. As grasping is complicated, we divide it into different stages and design a multi-stage reward function: first reaching the object, then closing the jaws, then picking up the object, and finally retracting the robotic arm to a specified position. 

% DR主要是为了弥合Sim2Real gap的嘛？我以为这个是为了抓取时的空间泛化性（我不懂我就问问）以及我看见总结句在最后，感觉挪到第一句会更清楚一点mxy ycy:dr两个效果都有，弥补Sim2Real gap和增强泛化性
\textbf{Domain Randomization.} 
% To mitigate the Sim2Real gap and improve the generalization in first-person view grasping, we adopt several domain randomization techniques.
To improve the generalization of the teacher policy and enhance the spatial diversity of the generated data, we adopt several domain randomization techniques:
\vspace{-0.7em}
\begin{itemize}
    \item \emph{objective position} $(x, y, z)$: To improve the spatial generalization of the teacher policy to the whole space, we apply the domain randomization for all three dimensions of \emph{x}, $y$, and $z$, ensuring that the $x$-coordinate and $y$-coordinate of the object 
    %, which is chosen randomly, 
    can cover a large range. Moreover, the height of the platform is randomized to ensure that the $z$-coordinate of the object in the data can cover the feasible working space of the robotic arm.
    \item \emph{object size}: To ensure that the agent can manipulate objects of \emph{various sizes}, we also randomize the size of the object.
\end{itemize}
\vspace{-0.7em}
These two domain randomization techniques allow the teacher policy to manipulate different objects of different positions and sizes, which is significant for real-world spatial and object generalization.

\subsection{Simulation Data Generation}
\label{sec_data_collect}
% After training the teacher policy, we utilize it to collect appropriate simulation data to train a student policy that can be deployed in the real world with generalization ability across different objects, grasping spatial positions, and backgrounds. 
% The data collection is fully automatic in Isaac Lab. %The system will automatically record action output and corresponding visual information during the inference of teacher policy.

After training the teacher policy, it can automatically generate data for training the student policy that can be deployed in the real world across different objects, spatial positions, and backgrounds.
% \xxz{We employ the teacher policy to automatically generate training data for a student policy that can be deployed in real-world settings across various objects, spatial grasping positions, and backgrounds.}
% For efficiency's sake, before generation, we need to determine the amount of data needed and decide the form of the data we want to collect and use.

\textbf{Data Volume Estimation.} To guide the data volume needed for policy generalization to a specific spatial volume, we analyze the sample efficiency for generalizing in a $b\times b\times b$ cube.
Based on the classical learning theory, the sample complexity of the uniform convergence property is in proportion to the VC dimension of the model hypothesis set~\citep{shalev2014understanding}. Extending previous results about generalization bounds for multi-task learning~\citep{crammer2012learning}, we can prove:
\begin{lemma}[Details are in Appendix~\ref{app_proof1}]
\label{lemma1}
For training a manipulating agent that can generalize spatially within a cube whose size is $b\times b\times b$ via imitation learning, under some mild assumptions in~\cite{crammer2012learning}, the VC dimension is at most proportional to $d\log \left(C b^3 d\right)$, where $d$ is the VC dimension of the hypothesis set for manipulating the fixed point and $C$ is a constant independent of $d, b$.
\end{lemma}

% This lemma demonstrates that \thk{only log-complexity growth of data}
Lemma~\ref{lemma1} demonstrates that the data required for policy generalization to a certain volume grows.
Consequently, training a student policy feasible under full spatial range requires a large scale of trajectories, which needs efficient data generation and collection.

On the other hand, as we cannot access the privileged information in the real world, the student policy needs visual observations that implicitly encode object position information. 
However, the Sim2Real gap between visual images in the simulator and the real world is significant due to the vast complexity of real-world environments (e.g., various backgrounds and objects).
% Additionally, vision-based policies are difficult to train on a large scale and require data augmentation.
To address this Sim2Real gap and enable robust spatial generalization, we propose using an individual vision module to extract the bounding box from camera views as a \textbf{consistent, low-dimensional, and complete} representation of the object’s spatial properties, such as position and size.

\textbf{Bounding Boxes Provides Sufficient 3D Information of Objects.} 
To efficiently generate a large amount of simulator data that aligns with real-world visual images, the vision module must be robust and efficient. 
% This requires the vision module to be pre-trained on a wide range of 2D internet image data, as 2D data is far more abundant than 3D data. 
Considering open-vocabulary detection, we selected YOLO-World~\citep{cheng2024yolo} as the frozen 2D visual feature extraction module.
% 这个frozen是什么意思呀？mxy。就是不训练，权重freeze住不动。只推理
% To generalize the  policy over different backgrounds and objects, 
% we consider using a frozen vision module to extract the visual features of the objects, which are then fed to the control policy. 
% \thk{why 2D bbox? 2D visual features of multiple viewpoints can imply 3D spatial information}
This provides us with the bounding box of the objects as a 2D low-dimensional visual feature, which is consistent and low-dimensional.
% In detail, we choose the bounding box as the 2D low-dimensional visual feature, and we use YOLO-World to extract the bounding boxes of the objects, which are then fed to the control policy. 
%because of its powerful open-vocabulary detection and generalization ability.
Notably, the inference time of YOLO-World for a single image is around 30ms, ensuring the efficiency of real-world inference.

Moreover, although 2D low-dimensional visual features may lose several information in images, we demonstrate that they are sufficient for manipulation as they can contain the full complete information about object position and size by presenting the following lemma, i.e., the 3D information of a sphere can be uniquely determined by the bounding boxes from two cameras:
\begin{lemma}[Details and proofs are in Appendix~\ref{app_proof2}]
\label{lemma2}
Given: (1) Two pinhole cameras with known intrinsic parameters \( K_i \) and extrinsic parameters \( (R_i, t_i) \); (2) A sphere of unknown radius $r$ is fully visible in both camera images without occlusion; (3) Normalized image coordinates of the bounding boxes of the sphere in both images: Camera $i$: \( (u_{1\text{min}}, v_{i\text{min}}, u_{i\text{max}}, v_{i\text{max}}) \), the image dimensions are known (width \( W_i \) and height \( H_i \) for each camera).
% \begin{itemize}
%     \item Two pinhole cameras with known intrinsic parameters \( K_i \) and extrinsic parameters \( (R_i, t_i) \).
%     \item A sphere of unknown radius $r$ is fully visible in both camera images without occlusion.
%     \item Normalized image coordinates of the bounding boxes of the sphere in both images: Camera $i$: \( (u_{1\text{min}}, v_{i\text{min}}, u_{i\text{max}}, v_{i\text{max}}) \), 
%     % Camera 2: \( (u_{2\text{min}}, v_{2\text{min}}, u_{2\text{max}}, v_{2\text{max}}) \). 
%     The image dimensions are known (width \( W_i \) and height \( H_i \) for each camera).
% \end{itemize}

Then the \textbf{3D coordinates of the sphere's center \( \mathbf{C} \) and radius \( r \) in the world coordinate system can be uniquely determined} with the known camera parameters and the normalized bounding box coordinates, provided that the cameras are not in degenerate configurations.
\end{lemma}

Lemma~\ref{lemma2} illustrates that the 3D position information of a sphere can be implied by utilizing multi-view bounding boxes~\citep{hartley2003multiple}, indicating that ManiBox can capture complete spatial position information of target objects for real-world control policies. 

In short, by using an individual vision module, Manibox mitigates the Sim2Real gap by providing the student policy with consistent input across both the simulation and the real world. Specifically, the student policy receives both proprioceptive data $\vo^{\text{rob}}_t$ and bounding box data $\vo^{\text{vis}}_t$ as inputs:
\begin{equation}
    \pi_{\theta}: \mathcal{O} = \mathcal{O}_{\text{rob}} \times \mathcal{O}_{\text{vis}} \rightarrow \mathcal{A}.
\end{equation}

\subsection{Student Policy Distillation}
\label{sec_student}

After collecting expert data from the teacher policy in the simulator, we distill the student policy with this expert data.
Then, we can directly deploy the student policy on the robot in the real world, which poses extra challenges compared with the simulation, such as sensor noise and dynamic environments. 

For example, open-vocabulary detection models may fail to recognize real-world target objects at times because of background distraction or robot gripper occlusion, while the recognition of the simulated target object rarely fails.
To address this challenge, we introduce the random mask. In detail, when distilling the student policy with the simulation expert data $\mathcal{D}$, we randomly mask some bboxes (i.e., set to 0) with a predefined probability of \texttt{random\_mask\_ratio}, indicating that this camera may not recognize the target object and imitating the real-world scenarios in which the open-vocabulary detection model fails to generate the bounding box.

Moreover, based on the real-world partial observability, our student policy requires encoding sufficient history information~\citep{hafner2019dream,lee2020context,ying2023task} and thus utilizes RNN as:
% \begin{equation}
% \begin{aligned}
%     \vx_t, \vh_t &= \mathrm{LSTM}(\vs^\prime_t, \vh_{t-1}),\\
%     \va_t &= \mathrm{FC_2}(\mathrm{drop}(\mathrm{gelu}(\mathrm{FC_1}(\vx_t)))),\\
% \end{aligned}
% \end{equation}
\begin{equation}
\vx_t, \vh_t = \mathrm{LSTM}(\vs^\prime_t, \vh_{t-1}),\quad 
\va_t = \mathrm{FC_2}(\mathrm{drop}(\mathrm{gelu}(\mathrm{FC_1}(\vx_t)))).
\end{equation}

where $\vs^\prime_t$ and $\vh_t$ denote the input state and the hidden state of the LSTM at time \( t \), $\vx_t$ is the intermediate output of the LSTM, and $\va_t$ is the action vector output by the policy. Furthermore, $\mathrm{LSTM}(\cdot), \mathrm{gelu}(\cdot),\mathrm{drop}(\cdot), \mathrm{FC_{\cdot}}(\cdot)$ represent the Long Short-Term Memory (LSTM)~\citep{HochSchm97}, Gaussian Error Linear Unit (GELU) activation~\citep{hendrycks2016gaussian}, dropout~\citep{srivastava2014dropout}, and Fully Connected (FC) layer, respectively.

The combination of random mask and trajectory modeling significantly contributes to robustness and generalization in deploying our student policy, conquering the challenge of partial observations and unseen real-world variations. (More Sim2Real techniques are in Appendix~\ref{Sim2Real})
% \revise{(Details of ablation study in Appendix~\ref{sec:ablation})}
% }

\section{Experiments}
\label{sec:exp}
In this section, we conduct extensive experiments in both the simulator and real world, covering various backgrounds, objects, and positions to answer the following questions: (1) What is the scaling relationship between spatial generalization and data amounts in manipulation? (Sec.~\ref{sec_scaling_law}) and (2) What about the efficacy of Manibox in real-world applications, especially the generalization ability to different backgrounds, objects, and object spatial positions? (Sec.~\ref{sec_real_world})
% \begin{itemize}
%     \item What is the scaling relationship between spatial generalization and data amounts in the grasping task? (Sec.~\ref{sec_scaling_law})
%     \item What about the efficacy of our method, especially its generalization ability to different backgrounds, objects, and object spatial positions in real-world applications? (Sec.~\ref{sec_real_world})
% \end{itemize}
% Through our experiments, we aim to justify (1) the efficacy of our method, its ability to generalize to different backgrounds, object variations, and object spatial positions 
% and (2) the scaling relationship between spatial generalization and data amounts in simple manipulation task.  \thk{the order of (1) and (2)?}

\begin{figure}[t]
\centering
\begin{subfigure}[b]{0.48\textwidth}
        \includegraphics[width=\linewidth]{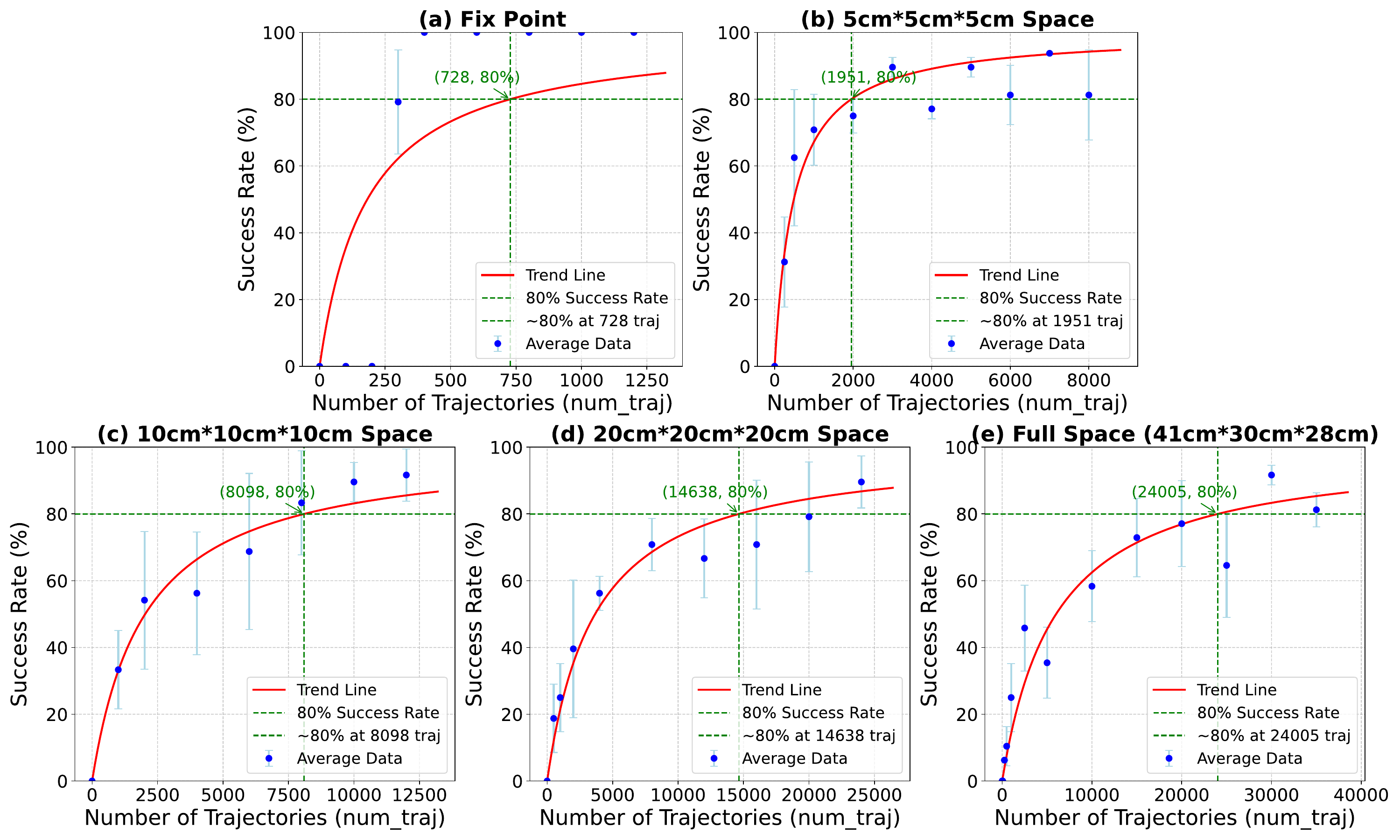}
        \caption{}
        \label{fig:success_vs_data}
    \end{subfigure}
\hfill
\begin{subfigure}[b]{0.48\textwidth}
        \includegraphics[width=\linewidth]{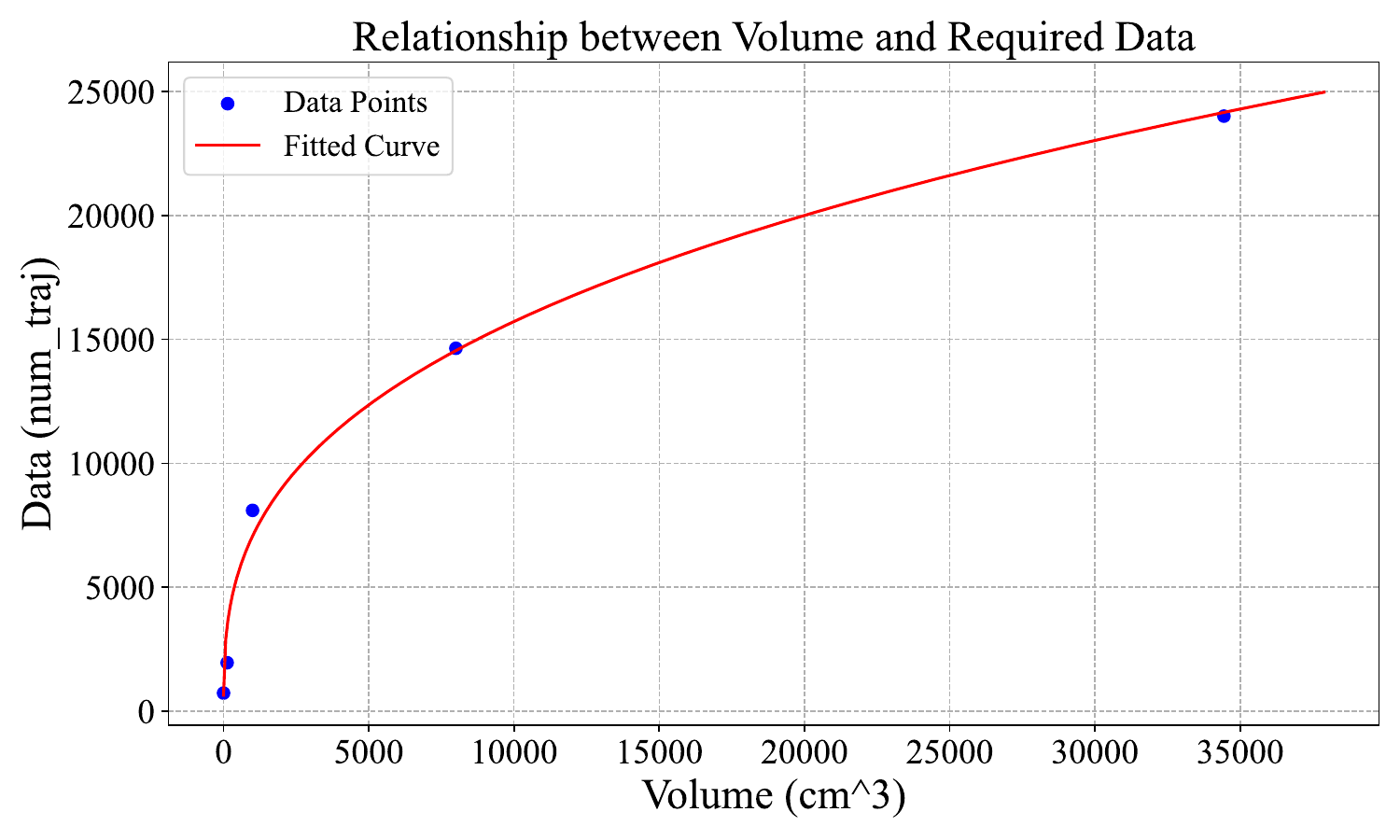}
        \caption{}
        \label{fig:data_vs_volume}
\end{subfigure}
\vspace{-0.5em}
\caption{(a) \textbf{The scaling relationship between spatial generalization and data volume in grasping}, under different spatial ranges. Data volume represents the number of trajectories used to train the student policy. The estimated 80\% success point is marked, with blue points showing the average success rate across 3 seeds and error bars indicating standard deviation. The curve corresponding to the other tasks is detailed in Appendix \ref{sec:pour_scaling_law}. (b) The relationship between spatial volume and data needed to reach 80\% grasping success rate. The fitted curve represents a power function $y=640.32\cdot x^{0.35} $.}
\label{fig:main}
\vspace{-0.25cm}
\end{figure}

\begin{table}[t]
\centering % Use \centering for simpler alignment
\footnotesize
\begin{tabular}{lccccc}
\toprule
\textbf{Method} &  Inverse Kinematics & Key Steps & w/r Random Mask & ManiBox (Ours)\\
\midrule
\textbf{SR of Full Space}   & $68.75\% \pm 5.10\%$
& $6.25\% \pm 8.84\%$
 & $58.33\% \pm 7.80\%$
 & \textbf{\boldmath $91.67\% \pm 2.95\%$}  \\
\bottomrule
\end{tabular}
\caption{The simulator success rate (SR) on the full space.}
\vspace{-0.75cm}
\label{tab:more_baseline}
\end{table}

% \begin{table}[t]
% \centering % Use \centering for simpler alignment
% \caption{The simulator success rate (SR) on the full space.}
% % \vspace{-0.5em}
% \begin{small}
% \begin{tabular}{lc}
% \toprule
% \textbf{Method} & \textbf{SR of Full Space} \\
% \midrule
% Inverse Kinematics (IK)  & \textbf{$68.75\% \pm 5.10\%$} \\
% Key Steps& \textbf{$6.25\% \pm 8.84\%$} \\
% Without Random Mask & \textbf{$58.33\% \pm 7.80\%$} \\
% \textbf{Student Policy of ManiBox (Ours)} & \textbf{\boldmath $91.67\% \pm 2.95\%$}  \\
% \bottomrule
% \end{tabular}
% \end{small}
% \label{tab:more_baseline}
% \end{table}

\subsection{Experimental Setup}
% \mxy{Important setting that should be specified here.}
% As a preliminary, we test our method on the grasping task

\textbf{Real Robot.} Mobile ALOHA~\citep{fu2024mobile} is a cost-effective and highly stable robot designed for manipulation tasks. 
Each subordinate arm of Mobile ALOHA is fitted with an RGB camera on its wrist, complemented by another RGB camera centrally mounted on the robot’s body. 
These three RGB images are taken as observations.
Mobile ALOHA has demonstrated strong generalization capabilities across a diverse range of complex manipulation tasks, making it an ideal choice for deploying Manibox.

\textbf{Tasks.} 
We choose several manipulation tasks to validate the generalization of Manibox. 
For each task, such as grasping an apple or pouring water from a bottle into a cup, we randomly place the object on a table on a spatial volume $41cm*30cm*28cm$ (length * width * height), which is roughly the maximum reach of our robotic arm and referred as "\textbf{Full Space}".
% First, we design a manipulation task in the simulator, i.e., 
% the robot arms are required to manipulate objects randomly placed on a table on a spatial volume $41cm*30cm*28cm$ (length * width * height), which is referred to as "Full Space", such as grasping an apple or pouring water from a bottle into a cup.
Also, to demonstrate the generalizability, real-world counterparts are established with alternative objects and backgrounds. 
% In both scenarios, three RGB images taken by wrist cameras and the exterior camera are taken as observations.

% We test our method in both the simulator and the real world to demonstrate its practicability, and justify the scaling law of data in simple manipulation tasks.

% \paragraph{Simulation Environment.}
%To minimize Sim2Real gap\xxz{more specific, from the aspects of high accuracy, real simulation, parallel training acceleration}, we choose Isaac Lab~\cite{} as the simulation environment. 
% We train our teacher policy in the simulator with PPO~\citep{schulman2017proximal} in \xxz{?} parallel environments with single GPU, taking privileged information and robot ontology information as input. Then, we use the trained teacher policy to inference in the simulator to collect position of robot arm, action and bounding box of object for each successful trajectory. 
% To maximize the speed of conducting experiments, we choose the highly parallelized Isaac Lab~\citep{mittal2023orbit} as the simulation environment.
% This allows us to conduct large-scale experiments on the relationship between success rate and data volume, as real-world experiments are time-consuming.

% \mxy{For abc reason}, we choose Isaacgym~\citep{makoviychuk2021isaac}, Isaac lab and Orbits~\citep{mittal2023orbit} as the simulation environment.

% Baseline放到ablation study讲吧？mxy
% \paragraph{Baselines.} ACT~\citep{zhao2023learning,fu2024mobile}
% \paragraph{Metrics} success rate

% \vspace{-0.5em}
\subsection{Spatial Scaling Law with Data}
\label{sec_scaling_law}

% \begin{figure}[h]
% \begin{center}
% \includegraphics[width=0.5\linewidth]{figs/volume_vs_data.pdf}
% %\framebox[4.0in]{$\; $}
% % \fbox{\rule[-.5cm]{0cm}{6cm} \rule[-.5cm]{12cm}{0cm}}
% \end{center}
% \vspace{-0.5cm}
% \caption{The relationship between spatial volume and data amounts needed to reach 80\% success rate in the grasping task.
% The data amounts are inferred from Fig.~\ref{fig:success_vs_data}.
% The x-axis represents the spatial volume. % in power rate relationship.
% The y-axis represents the data amounts needed, namely the number of trajectories used for training the student policy.
% The fitted curve represents a power function of $y=1507\cdot x^{0.28} $.
% % The fitted curve represents a logarithmic function of \textit{a\cdot \ln(b \cdot x) + c}.
% %\[a \cdot \ln(b \cdot x) + c\] 
% % \thk{a * np.log(b * x) + c 
% % Fitted parameters:
% % a = 2449.38
% % b = 0.39
% % c = -909.57} 
% }
% \vspace{-0.3cm}
% \label{fig:data_vs_volume}
% \end{figure}

To guide us in determining the data volume required for a certain level of spatial generalization, we investigate the relationship between policy performance and data volume.
% To investigate the relationship between data volume and policy performance, 
We conduct extensive experiments in the simulator to evaluate the success rate (\textbf{SR}) of policies trained on different data volumes.
We select five distinct spatial ranges to explore this scaling behavior, including the fixed range and four dynamic ranges of 5, 10, 20 units, and full space. 
% For each range, we select more than 8 \xxz{Unified standards?}different data quantities to evaluate the success rate of the trained policies in the simulator. 
For each spatial range, we select more than 8 data volumes to train the student policy and evaluate its grasping SR at random positions.

Fig.~\ref{fig:success_vs_data} shows the \textbf{Michaelis-Menten kinetic curve} between the SR and the volume of the data:
when the data volume is zero for each spatial range, the policy inevitably fails;
as the data volume increases, the manipulation policy's SR improves, but the growth slows down. Eventually, the SR tends to be \textbf{saturated} because of the regret problem and is infinitely close to 100\%. 
The relationship between the SR and the data volume is \textbf{expressed by the formula: $SR = {100\% * D \over K_m + D}$}, where $D$ is the data volume of the imitation policy, $K_m$ is the data volume required to achieve 50\% SR.

% surpasses 80\% and tends to 100\%.

Furthermore, we identify the data volume required to achieve an 80\% SR for each spatial range (Fig.~\ref{fig:success_vs_data}) and find that the spatial volume and the required data follow a \textbf{power law relationship} (Fig.~\ref{fig:data_vs_volume}). 
% The data needed to reach high SRs increases linearly with the \textit{log} of the generalized space volume. \thk{FIX: positive correlation}
% This further validates Lemma~\ref{lemma1} introduced above and provides guidance for data-driven spatial generalization:
% This power law provides guidance for data-driven spatial generalization:
% The more data we have, the higher the robot's manipulation SR. 
In other words, the data volume required for spatial generalization is related to the spatial volume in a power-law relationship, i.e., more data improves the ability to generalize over a larger spatial range.
For example, to achieve spatial generalization across an $x$-fold increase in volume, the required data volume scales approximately as $x^{0.35}$.
% in our settings, if we want to generalize to $x$ times the spatial volume, then the data volume needs to be \textbf{expanded by about $x^{0.35}$ times}.
In our experimental setup, generalizing from $1\text{cm}^3$ to $34400\text{cm}^3$ necessitates a $34400^{0.35} \approx 38$-fold increase in data volume.
% In the setting in the paper, $34400cm^3$ versus $1cm^3$, the data volume required for $80\%$-SR spatial generalization of the former is $34400^{0.35}=38$ times that of the latter.
% As the spatial range increases, more data is required, and the data volume needed grows with the space volume as a power rate relationship.
% In fact, for the same task, achieving high generalization success over a $41 cm * 30cm * 28cm = 34440 cm^3$ range may require at least 20,000 trajectory data from different spatial positions. 
Such large-scale data collection would be expensive in the real world, emphasizing the significance of simulation data, which is more affordable and efficient.
% Furthermore, we observe that larger positional ranges require more data to achieve comparable levels of generalization. This suggests that data efficiency varies with the complexity of the positional range, reinforcing the importance of optimizing data usage to balance training effort and model performance.
To further demonstrate the importance of other variables (e.g., time) in spatial intelligence and the robustness of policy generalization to visual uncertainty, 
we have provided detailed ablation experiments on the spatial intelligence of Manibox in Appendix~\ref{sec:ablation}.
\subsection{Real-world Results}
\label{sec_real_world}

\begin{figure}[t] 
\begin{center}
\includegraphics[width=0.8\linewidth]{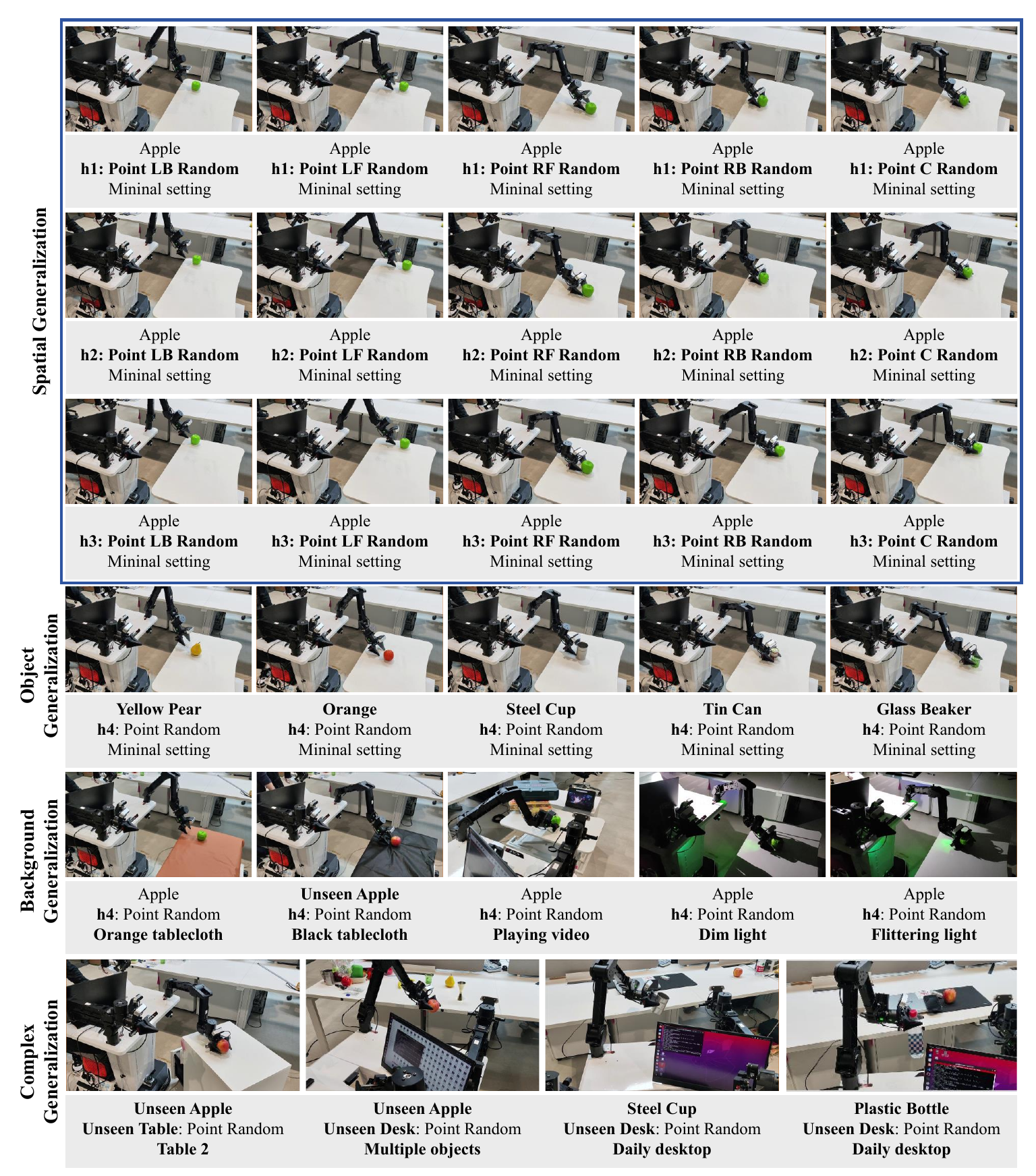}
%\framebox[4.0in]{$\; $}
% \fbox{\rule[-.5cm]{0cm}{6cm} \rule[-.5cm]{12cm}{0cm}}
\end{center}
\vspace{-0.5cm}
\caption{Demonstration of the generalization of Manibox across different backgrounds, objects, and positions.
The h1, h2, h3, and h4 are approximately 57cm, 65cm, 72cm, and 65cm, which are randomly selected object heights based on the height of our robot.
LB, LF, RF, RB, and C are abbreviations for Left-Back, Left-Front, Right-Front, Right-Back, and Center area, respectively.
}
\vspace{-1.5em}
\label{fig:demo_fig}
\end{figure}

To validate the generalization ability of Manibox in the real world, we deploy student policies trained on simulation data into a real robot and test the performance across multiple arbitrarily chosen positions, eight objects (some are out-of-distribution), five backgrounds, and three complex settings, as illustrated in Figure~\ref{fig:demo_fig}. 
% More tasks, pictures, and videos of the experiment are available in the Appendix~\ref{sec:more_exp} and supplementary materials.
More details of these experiments are in the Appendix~\ref{app:real_eval}.

% To verify that our approach maintains good spatiazl, object, and background generalization in real-world scenarios, we deploy models trained on simulation data onto a real robot and test their performance on grasping task across multiple arbitrarily chosen positions, eight objects(with some of them out-of-distribution), five backgrounds and three unseen settings as illustrated in Figure~\ref{fig:demo_fig}.

% \begin{figure}[t]
% \begin{center}
% %\framebox[4.0in]{$\; $}
% \fbox{\rule[-.5cm]{0cm}{6cm} \rule[-.5cm]{12cm}{0cm}}
% \end{center}
% \caption{Generalization on various scenarios, objects, and position of objects}
% \end{figure}

% \paragraph{Position generalization.}To confirm the feasibility of our method, we assess the generalization capability of our student policy across different positional ranges. Same as simulator, we deployed five distinct configurations corresponding to fixed points and dynamic ranges of 5, 10, 20, and 30 units. These strategies were evaluated on real hardware to measure their SRs, utilizing 10 randomly chosen points for each deployment.
% \vspace{-0.7em}
% \paragraph{Spatial generalization.} 
\textbf{Spatial generalization.} 
To evaluate the generalization capability of Manibox, we evaluate the student policy across four spatial volumes and a fixed point. 
% We deploy the student policy trained on the data of full space 
% Based on the scaling curve between data volume and SR, we select the optimal policy in the simulator for validation on the real hardware. 
% These policies were then evaluated on real hardware to measure their SRs, utilizing 10 randomly chosen points for each deployment.
% First, student policy is trained with simulator data which cover the expected spatial range and deployed on the real robot. 
For each experiment, we randomly place an apple on the table at arbitrarily chosen heights within the given range(as illustrated in the first three rows of Fig.~\ref{fig:demo_fig}).
% The process is then repeated to test the SR of the grasping policy. 
Alongside the difference in form and appearance, the apple presents considerable unpredictability to the visual detection model due to factors including background, lighting, target object, etc. 
Consequently, the bounding box varies in size or even disappears at most of the time, emphasizing the need for robustness of the policy.
As shown in Tab.~\ref{real_success_table} (Appendix~\ref{app_real_world_exp}), Manibox generalizes well to objects falling in different spatial volumes, achieving the SR between 70\%-100\%.
In contrast, vision-based ACT, which trains the student policy with the vision input ACT~\citep{zhao2023learning} on video-action simulation data, experiences a visible drop in SR when spatial range scales up due to the Sim2Real gap, despite its considerable SR of 70\% on a fixed point.
% The desired performance of the baseline further demonstrates the significance and necessity of exploring spatial generalization.
The results demonstrate that Manibox harbors strong generalization when addressing uncertainties arising from real-world objects, variations in spatial positions, and visual detection interferences.

% \xxz{data needed, table to record SR, what's the format?}

% To ensure alignment between simulation and real-world performance, we initially evaluated a broader set of strategies within Isaac Lab. The top-performing strategy from the simulator is selected for real hardware deployment.

% \begin{figure}[t]
% \begin{center}
% \includegraphics[width=0.8\linewidth]{figs/demo_pour_merge.jpg}
% %\framebox[4.0in]{$\; $}
% % \fbox{\rule[-.5cm]{0cm}{6cm} \rule[-.5cm]{12cm}{0cm}}
% \end{center}
% \vspace{-0.2cm}
% \caption{\textbf{Pour Water.} 
% The robot grasps the left water bottle and pours water into the right cup, demonstrating the \textbf{extensibility of Manibox to more complex tasks}.}
% \vspace{-0.75em}
% \label{pour_water}
% \end{figure}

% \begin{figure}[t]
% \begin{center}
% \includegraphics[width=0.8\linewidth]{figs/demo_handle.jpg}
% %\framebox[4.0in]{$\; $}
% % \fbox{\rule[-.5cm]{0cm}{6cm} \rule[-.5cm]{12cm}{0cm}}
% \end{center}
% \vspace{-0.2cm}
% \caption{\textbf{Grab the handle of the cup}. Manibox is flexible to combine with more fine-grained visual models like~\cite{liu2023grounding} to achieve more complicated manipulation tasks. 
% We take grabbing the handle of a cup as an example. 
% The first figure shows the recognition of the cup handle from the first-person view, and other figures show the task completion process from the third-person view.}
% \vspace{-1.25em}
% \label{grasp_cup_handle}
% \end{figure}

\begin{figure}[t]
\begin{center}
\includegraphics[width=0.8\linewidth]{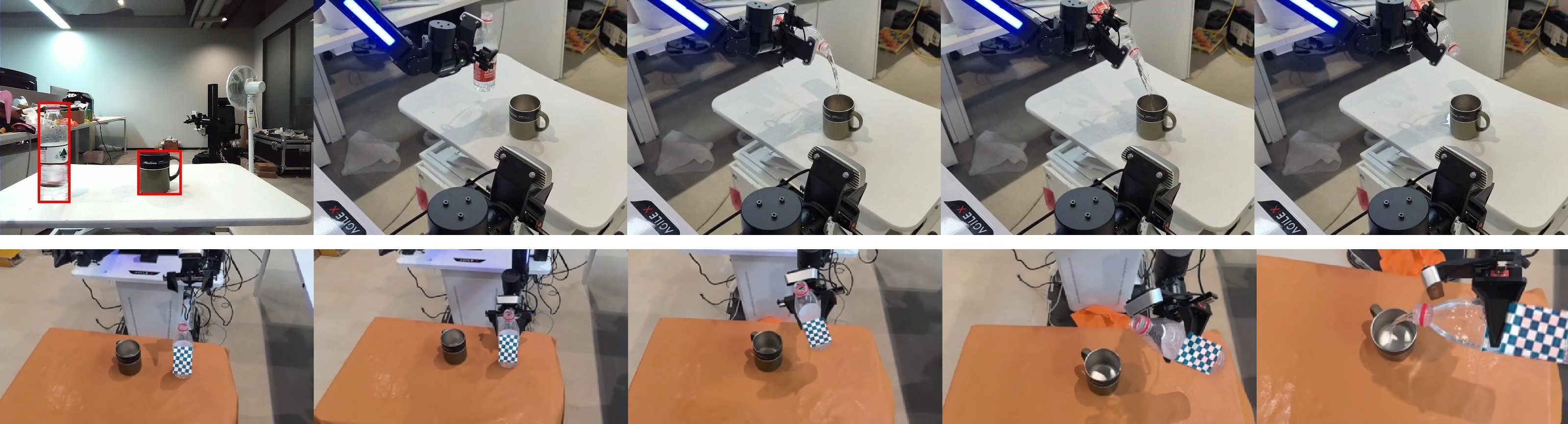}
\includegraphics[width=0.8\linewidth]{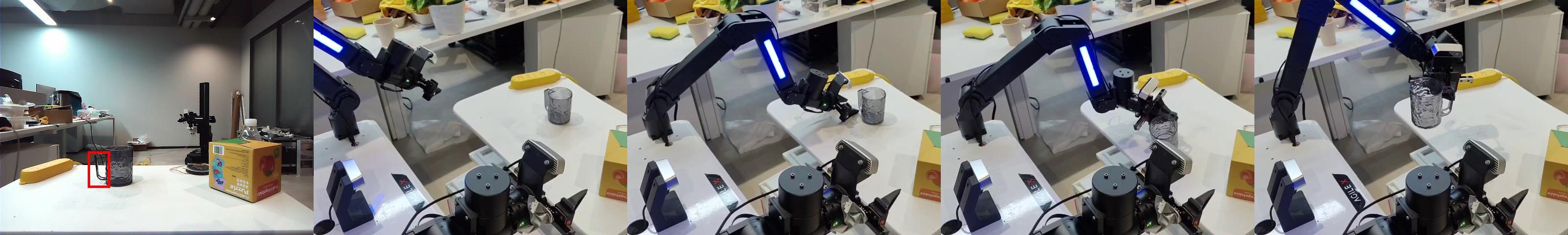}
%\framebox[4.0in]{$\; $}
% \fbox{\rule[-.5cm]{0cm}{6cm} \rule[-.5cm]{12cm}{0cm}}
\end{center}
\vspace{-0.2cm}
\caption{
\textbf{Pour Water (the above two rows)}:
The robot grasps the left water bottle and pours water into the right cup, demonstrating the extensibility of Manibox to more complex tasks.
\textbf{Grab the handle of the cup (the last row)}: Manibox is flexible to combine with more fine-grained visual models like~\cite{liu2023grounding} to achieve more complicated manipulation tasks. 
We take grabbing the handle of a cup as an example. 
The first figure shows the recognition of the cup handle from the first-person view, and other figures show the task completion process from the third-person view.}
\vspace{-1.5em}
\label{pour_water_grasp_cup_handle}
\end{figure}

% \vspace{-0.7em}
% \paragraph{Object generalization.} 
\textbf{Object generalization.} 
To justify Manibox's object generalization ability, we chose eight objects: the original apple, unseen apple, yellow pear, orange, steel cup, tin can, glass beaker, and plastic bottle. 
For each object, the open-vocabulary visual detection model and obtain the bounding box from the corresponding name, which is enough for the policy to grasp the target object.
% Tab.~\ref{tab:experiment_settings} shows detailed object size information for training and inference.
It is worth noting that five of the selected objects are out-of-distribution in size (their bounding box sizes are unseen in the simulation data), which further illustrates the object generalization ability of Manibox.  
The ability to generalize across objects, even with significantly different bounding boxes, aligns with our Lemma~\ref{lemma2}, which proves that the 3D position information of the object is complete in the dataset.
% This is likely because our Lemma~\ref{lemma2} ensures the 3D position information of the object is complete in the dataset, allowing for a certain level of generalization even when there is significant variation in the bounding box.

% For each object, we selected \mxy{x} points in \mxy{x} spatial range. \mxy{TODO: conclusion and analysis}. 

% \paragraph{Background Generalization.}
\textbf{Background Generalization.}
The open-vocab perception models (e.g. YOLO-World) and learning from trajectories enable ManiBox to generalize across vision backgrounds without extra visual domain randomization.
To verify it, we consider three table shapes, two tablecloths colors, as well as three lighting conditions: a single stationary point light source, a single flickering point light, and playing videos in the background. 
Furthermore, we construct two more complex scenarios: a multi-object environment and a typical daily workplace desktop.
The results demonstrate Manibox's strong background generalization ability. 
Even in dark environments, the detection model fails to detect the target object most of the time,
\textbf{the generalization of Manibox compensates for those visual errors} (Sec. \ref{sec:random_mask_inference}). 
This can be attributed to Manibox's endeavor to integrate historical information and random masking to increase the visual model’s tolerance for errors. Details are in Appendix~\ref{sec:ablation}.
% ~\ref{sec:key_steps}-\ref{sec:random_mask_inference}.}

% \paragraph{Various manipulation tasks.} 
\textbf{Various manipulation tasks.} 
In Figure~\ref{pour_water_grasp_cup_handle}, we also evaluate more real-world manipulation tasks: \textbf{Pour Water} and \textbf{Grab the handle of the cup}, showcasing the extensibility of Manibox to multi-object manipulation tasks and the grasping of detailed parts of irregular objects, respectively.

\subsection{Baseline Comparison}
\label{sec:baselines}
We further evaluate ManiBox on multiple manipulation tasks of RoboTwin~\cite{chen2025robotwin}.
Specifically, we select representative tasks spanning bimanual and object-centric scenarios.
All policies are trained on the Aloha-AgileX using 50 \texttt{demo\_clean} demonstrations for each task. 
All models are evaluated 100 times under \texttt{demo\_randomized} (Hard) settings, following the official RoboTwin 2.0 protocol.
Overall, ManiBox matches or surpasses ACT, DP, and DP3, with clear gains in the challenging \texttt{Hard} settings (Table~\ref{tab:baseline}).

\begin{table}[t]
    \centering
    \caption{Comparison with baselines on RoboTwin tasks (\texttt{Hard} Setting)}
    \label{tab:baseline}
    \resizebox{0.5\linewidth}{!}{
    \begin{tabular}{lcccc}
        \toprule
        \textbf{Task} & ACT & DP & DP3 & \textbf{Ours} \\
        \midrule
        Move Can Pot & 0\% & 0\% & 6\% & \textbf{11\%} \\
        Grab Roller  & 25\% & 0\% & 2\% & \textbf{43\%} \\
        Stack Bowls Two & 0\% & 0\% & \textbf{6\%} & 1\% \\
        Pick Dual Bottles & 0\% & 0\% & \textbf{1\%} & \textbf{1\%} \\
        % Place Dual Shoes & 0\% & 0\% & 0\% & 0\% \\
        % Stack Blocks Two & 0\% & 0\% & 0\% & 0\% \\
        Put Bottles Dustbin & 1\% & 0\% & \textbf{21\%} & 1\% \\
        \midrule
        \textbf{Average Success Rate} & 5.2\% & 0\% & 7.2\% & \textbf{11.4\%} \\
        \bottomrule
    \end{tabular}}
\end{table}

\section{Conclusion}
In this work, we present ManiBox, a novel bounding-box-guided manipulation method to decouple Internet-scale-trained perception models from policy generalization whose action data are scalably collected in the simulator.
To reduce the generalization gap in vision-based models from simulator to real-world deployment, Manibox equips the state-based student policy with bounding-box as visual intermediate inputs. 
% This policy is trained on scalable simulation data generated by the teacher policy.
Extensive experiments in both simulated and real-world environments demonstrate that ManiBox significantly outperforms existing methods in adapting across different spatial positions, objects, and background variations. 
Moreover, our theoretical and empirical analyses reveal that the data volume required for spatial generalization is related to the spatial volume in a power-law relationship, and the SR follows Michaelis-Menten kinetics for specified spatial volumes.
These findings pave the way for the development of robust spatial generalization across a wide range of embodied systems.
More discussion of the limitations and future work is in Appendix~\ref{limitation}.

% \newpage
{
    \small
     \bibliographystyle{plain}
    \bibliography{main}
}

%%%%%%%%%%%%%%%%%%%%%%%%%%%%%%%%%%%%%%%%%%%%%%%%%%%%%%%%%%%%

\newpage
\appendix

\section{Proofs}

\subsection{Details and Proof of the Lemma~\ref{lemma1}}
\label{app_proof1}
We follow the setting in~\cite{crammer2012learning}. First, assume the hypothesis set of policies to grasp a fixed single point is $\mathcal{H}$ and its VC dimension is $d$. When considering $T$ tasks, each represented by a different grasping position, we take the $k$-shared task classifier $(\mathcal{H}_k, g)$~\citep{crammer2012learning}. Here $\mathcal{H}_k = \{h_1, ..., h_k\}\subseteq \mathcal{H}$ is a subset of $\mathcal{H}$ and $g$ maps each task $t\in\mathcal{T}$ to some $h_i\in\mathcal{H}_k$. Thus the $k$-shared task classifier hypothesis is defined as $\mathcal{F}_{\mathcal{H}, k} = \{f_{\mathcal{H}_k, g}: |\mathcal{H}_k| = k, \mathcal{H}_k\subseteq \mathcal{H}, g: \mathcal{T}\rightarrow K\}$, here $f_{\mathcal{H}_k, g}(\cdot, t) = h_{g(t)}(\cdot)$. Then~\cite{crammer2012learning} has shown that the VC dimension of $\mathcal{F}_{\mathcal{H}, k}$ is at most $\mathcal{O}(T\log k + kd \log(Tkd))$.

As we consider $\mathcal{H}$ as the set of neural networks, previous works have shown that the VC dimension of multi-layer neural networks is around the parameters of the neural networks~\citep{maass1994neural,sontag1998vc}. Thus $d \gg T$ in our setting and $\mathcal{O}(T\log k + kd \log(Tkd))\approx \mathcal{O}(kd \log(Tkd))$.

Finally, we consider the task number $T$. Assume that grasping two points of which the distance below $\epsilon$ is similar, we need around $b^3 / \frac{4}{3} \pi \epsilon^3$ points for covering the grasping cube with size $b\times b\times b$. Thus, the VC dimension is at most
\begin{equation}
\begin{split}
     \mathcal{O}(kd \log(Tkd)) = & \mathcal{O}(kd \log(\frac{b^3}{\frac{4}{3} \pi \epsilon^3}  kd)) \\
     = & \mathcal{O}(kd \log(\frac{3 b^3 kd}{4 \pi \epsilon^3} )).
\end{split}
\end{equation}

\subsection{Proof of the Lemma~\ref{lemma2}}
\label{app_proof2}

\begin{proof}

Firstly, we convert normalized coordinates to pixel coordinates:

\begin{align*}
u_{i\text{min(pixel)}} &= u_{i\text{min}} \times W_i \\
v_{i\text{min(pixel)}} &= v_{i\text{min}} \times H_i \\
u_{i\text{max(pixel)}} &= u_{i\text{max}} \times W_i \\
v_{i\text{max(pixel)}} &= v_{i\text{max}} \times H_i
\end{align*}

Then, we compute the center of the bounding box in pixel coordinates:
\begin{align*}
u_{i\text{center}} &= \frac{u_{i\text{min(pixel)}} + u_{i\text{max(pixel)}}}{2} \\
v_{i\text{center}} &= \frac{v_{i\text{min(pixel)}} + v_{i\text{max(pixel)}}}{2}
\end{align*}
These points represent the projections of the sphere's center onto the image planes.

% \textbf{Step 2: Back-Project Center Points to Viewing Rays}

Secondly, for each camera, we back-project the center point into a 3D viewing ray in camera coordinates:
\[
\mathbf{d}_i = K_i^{-1} \begin{bmatrix} u_{i\text{center}} \\ v_{i\text{center}} \\ 1 \end{bmatrix}
\]
This gives the direction vector \( \mathbf{d}_i \) in the camera coordinate system.

% \textbf{Step 3: Express Sphere Center in Camera Coordinates}

Therefore, we can express the sphere center in camera coordinates since the sphere's center lies somewhere along the viewing ray:
\[
\mathbf{C}_{i, \text{cam}} = \lambda_i \mathbf{d}_i
\]
where \( \lambda_i > 0 \) is an unknown scalar representing the distance along the ray from the camera center to the sphere's center.

% \textbf{Step 4: Transform Sphere Center to World Coordinates}

Thirdly, we transform the sphere center from camera coordinates to world coordinates:
\[
\mathbf{C} = R_i^\top \mathbf{C}_{i, \text{cam}} + \mathbf{t}_i = R_i^\top (\lambda_i \mathbf{d}_i) + \mathbf{t}_i
\]
This must hold true for both cameras:
\[
\mathbf{C} = R_1^\top (\lambda_1 \mathbf{d}_1) + \mathbf{t}_1 = R_2^\top (\lambda_2 \mathbf{d}_2) + \mathbf{t}_2
\]
% \textbf{Step 5: Set Up System of Equations}
By equating the expressions for \( \mathbf{C} \) we get:
\[
R_1^\top (\lambda_1 \mathbf{d}_1) + \mathbf{t}_1 = R_2^\top (\lambda_2 \mathbf{d}_2) + \mathbf{t}_2
\]
By rewriting we have:
\[
R_1^\top \mathbf{d}_1 \lambda_1 - R_2^\top \mathbf{d}_2 \lambda_2 = \mathbf{t}_2 - \mathbf{t}_1
\]
Let’s denote:
\begin{align*}
\mathbf{a}_1 &= R_1^\top \mathbf{d}_1 \\ 
\mathbf{a}_2 &= R_2^\top \mathbf{d}_2 \\ 
\mathbf{b} &= \mathbf{t}_2 - \mathbf{t}_1
\end{align*}
So the equation becomes:
\[
\mathbf{a}_1 \lambda_1 - \mathbf{a}_2 \lambda_2 = \mathbf{b}
\]
% \textbf{Step 6: Solve for \( \lambda_1 \) and \( \lambda_2 \)}
Note that this is a system of three linear equations with two unknowns \( \lambda_1 \) and \( \lambda_2 \):
\[
\begin{cases}
a_{1x} \lambda_1 - a_{2x} \lambda_2 = b_x \\
a_{1y} \lambda_1 - a_{2y} \lambda_2 = b_y \\
a_{1z} \lambda_1 - a_{2z} \lambda_2 = b_z
\end{cases}
\]
Since we have more equations than unknowns (overdetermined system), we can solve for \( \lambda_1 \) and \( \lambda_2 \) using least squares estimation.

% \textbf{Constructing the Linear System:}

% Let’s write it in matrix form:
Note that this can be written in matrix form:
\[
\begin{bmatrix}
a_{1x} & -a_{2x} \\
a_{1y} & -a_{2y} \\
a_{1z} & -a_{2z}
\end{bmatrix}
\begin{bmatrix}
\lambda_1 \\
\lambda_2
\end{bmatrix}
=
\begin{bmatrix}
b_x \\
b_y \\
b_z
\end{bmatrix}
\]
% \textbf{Solving the System:}
Let \( A \) be the \( 3 \times 2 \) matrix of coefficients, \( \boldsymbol{\lambda} \) the vector of unknowns, and \( \mathbf{b} \) the vector of constants:
\[
A = \begin{bmatrix}
a_{1x} & -a_{2x} \\
a_{1y} & -a_{2y} \\
a_{1z} & -a_{2z}
\end{bmatrix}, \quad \boldsymbol{\lambda} = \begin{bmatrix} \lambda_1 \\ \lambda_2 \end{bmatrix}, \quad \mathbf{b} = \begin{bmatrix} b_x \\ b_y \\ b_z \end{bmatrix}
\]
We can compute the least squares solution:
\[
\boldsymbol{\lambda} = (A^\top A)^{-1} A^\top \mathbf{b}
\]
% \textbf{Step 7: Compute the Sphere Center \( \mathbf{C} \)}
Once \( \lambda_1 \) and \( \lambda_2 \) are found, \textbf{\( \mathbf{C} \) can be calculated using either camera's equation}:
\[
\mathbf{C} = R_1^\top (\lambda_1 \mathbf{d}_1) + \mathbf{t}_1
\]
Alternatively, we compute both and average them for robustness:
\[
\mathbf{C} = \frac{1}{2} \left( R_1^\top (\lambda_1 \mathbf{d}_1) + \mathbf{t}_1 + R_2^\top (\lambda_2 \mathbf{d}_2) + \mathbf{t}_2 \right)
\]
% \textbf{Step 8: Compute the Radius \( r \) of the Sphere}
Finally, since \( \lambda_i \) and \( \lambda_2 \) are found, we can compute the distance \( Z_i \) from the camera along the viewing direction \( \mathbf{d}_i \):
\[
\mathbf{C}_{i, \text{cam}} =  \lambda_i \mathbf{d}_i = Z_i \cdot \mathbf{\hat{d}}_i
\]
where \( \mathbf{\hat{d}}_i \) is normalized by:
\[
\mathbf{\hat{d}}_i = \frac{\mathbf{d}_i}{\|\mathbf{d}_i\|}
\]
The projected radius of the sphere \( s_i \) (in pixels) is half the size of the bounding box width or height (assuming the sphere projects to a circle):
\[
s_i = \frac{w_i + h_i}{4}
\]
where 
\begin{align*}
w_i &= u_{i\text{max(pixel)}} - u_{i\text{min(pixel)}} \\
h_i &= v_{i\text{max(pixel)}} - v_{i\text{min(pixel)}}
\end{align*}
Consider similar triangles in the imaging geometry:
\[
\frac{s_i}{f_i} = \frac{r}{Z_i}
\]
% Solving for \( r \):
\textbf{We can compute the radius of the sphere}:
\[
r = \frac{s_i \cdot Z_i}{f_i}
\]
% \textbf{Step 9: Uniqueness of the Solution}
Additionally, the solution is unique provided that:
\begin{itemize}
    \item \textbf{The Cameras Are Not Aligned:} The direction vectors \( \mathbf{a}_1 \) and \( \mathbf{a}_2 \) are not colinear.
    \item \textbf{Non-Degenerate Configuration:} The cameras have different positions or orientations, ensuring the lines of sight intersect at a single point.
\end{itemize}

Under these conditions, the least squares solution yields a unique \( \boldsymbol{\lambda} \), and thus unique \( \mathbf{C} \) and \( r \).

% \textbf{Conclusion:}

% The 3D coordinates of the sphere's center \( \mathbf{C} \) and the radius \( r \) of the sphere can be uniquely determined using:
% \begin{itemize}
%     \item The intrinsic and extrinsic parameters of the cameras.
%     \item The bounding box centers from the images.
% \end{itemize}

% This is accomplished through triangulation based on the known camera geometries and the projection of the sphere's center in both images.

% ---

% \textbf{Final Statement:}

% Under the given conditions and assumptions, the sphere's center coordinates \( \mathbf{C} \) can be uniquely determined using the known intrinsic and extrinsic parameters of the two cameras and the normalized bounding box coordinates, \textbf{even when the sphere's radius \( r \) is unknown}. The solution involves back-projecting the bounding box centers into 3D rays, setting up a system of linear equations based on camera geometry, and solving for the distances along the rays to compute the sphere's center in the world coordinate system.

\end{proof}

\section{Ablation Study}
\label{sec:ablation}

% Spatial intelligence is the capability of machines to perceive, reason, and act within a three-dimensional space and over time, allowing them to understand the positions and interactions of objects in a 4D context (three spatial dimensions plus time). Through a series of ablation experiments, we demonstrated that both spatial and temporal information are essential components for achieving spatial intelligence.
To demonstrate the essential role of 3D spatial and 1D temporal information for spatial intelligence, we perform an additional experiment focusing on \textbf{temporal variability}, in addition to the spatial generalization tests in Sec.~\ref{sec:exp}. This experiment includes an ablation study comparing models trained on a few selected key timesteps with those trained on all timesteps in a full trajectory. 
To further illustrate the robust inference capabilities achieved by the enhanced \textbf{policy generalization}, we test the policy's ability to handle significant uncertainty in the output of the visual model. During inference, we apply random masks and noise to the visual model's bounding box outputs, allowing us to evaluate the policy's generalization ability under such conditions.

\subsection{Training with Only Key Steps.}
\label{sec:key_steps}
To assess the impact of temporal information on policy effectiveness, we conduct an experiment where the model is trained using only key steps from the trajectory. 
The specific key steps selected were [0, 18, 20, 22, 70], chosen to represent pivotal moments in the trajectory for the grasping task: 0 corresponds to the initial state before the gripper starts approaching the apple, 70 represents the final state when the robotic arm retracts, and steps 18 to 22 denote intermediate states where the gripper makes contact with the apple and closes around it.

\begin{table}[h]
\centering % Use \centering instead of \begin{center} ... \end{center} for simpler alignment
\caption{Success Rates for Training with Only Key Steps}
\begin{small}
\begin{tabular}{lcc}
\toprule
\textbf{Experiment} & \textbf{Real Robot SR} & \textbf{Simulation SR} \\
\midrule
Training with Full Trajectory & 90\% & 81.25\% \\
Training with Key Steps & 0\% & 4.17\% \\
\bottomrule
\end{tabular}
\end{small}
\label{tab:key_steps_results}
\end{table}

The results in Tab.~\ref{tab:key_steps_results} indicate a stark difference in performance between training with the full trajectory and training with only key steps. Training with the full trajectory resulted in a success rate of 90\% on the real robot and 81.25\% in simulation. In contrast, training with only key steps leads to complete failure on the real robot (0\% success rate) and a significantly reduced success rate of 4.17\% in simulation. This outcome highlights that relying solely on key steps fails to provide the temporal continuity necessary for effective spatial generalization. The policy's inability to capture sufficient context from isolated states results in poor action generation and overall task failure.

\subsection{Robustness against Detection Failure during Inference.}
\label{sec:random_mask_inference}
During the inference phase, visual models such as YOLO-World may not always successfully identify objects, leading to significant uncertainty in visual input. 

\paragraph{Our policy still works well in situations where the visual model fails to detect.}

% \mxy{TODO, Cases of failure to detect, especially in  flittering light. Our policy still works well in this situation.}

\begin{figure}[t]
\begin{center}
\includegraphics[width=\linewidth]{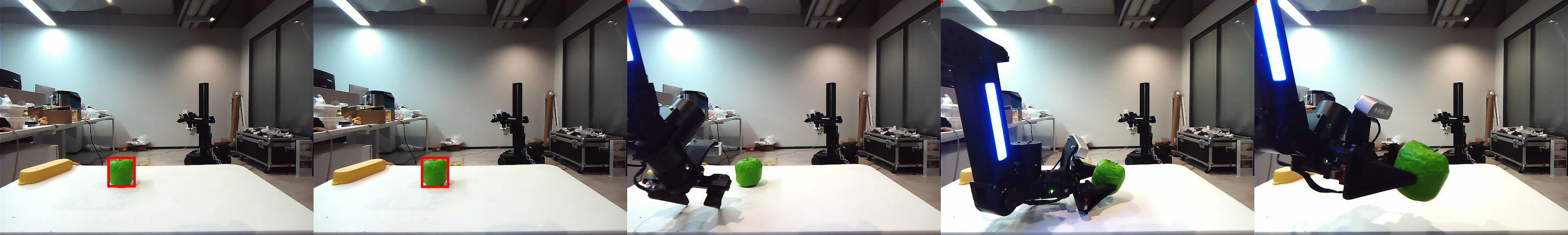}
%\framebox[4.0in]{$\; $}
% \fbox{\rule[-.5cm]{0cm}{6cm} \rule[-.5cm]{12cm}{0cm}}
\end{center}
\vspace{-0.3cm}
\caption{A case of detection failure. The task is to catch apples to a specified location in a normal environment.}
\label{failure_detection_1}
\end{figure}

\begin{figure}[t]
\begin{center}
\includegraphics[width=\linewidth]{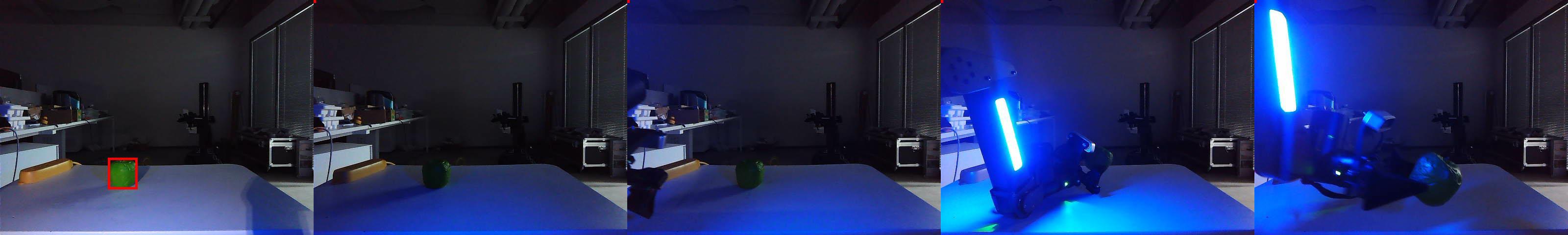}
%\framebox[4.0in]{$\; $}
% \fbox{\rule[-.5cm]{0cm}{6cm} \rule[-.5cm]{12cm}{0cm}}
\end{center}
\vspace{-0.3cm}
\caption{A case of detection failure. The task is to catch apples at a specified location in a flittering light environment.}
\label{failure_detection_2}
\end{figure}

For example, visual models may fail to detect target objects when the robotic arm approaches the target, under dim lighting conditions, or when there is partial occlusion of the target by the background (see Figures~\ref{failure_detection_1} and \ref{failure_detection_2}).
% \thk{TODO, ref the figure}
Especially, when using flickering and swaying point light sources in dimly lit environments, the detection by visual models is particularly unstable.
Such failures may occur at any timestep during manipulation and may persist for several timesteps.
Moreover, these failures are unstable and unpredictable. For the same object placement, the occurrence of detection failures during manipulation does not consistently happen or not happen at the same timestep.

Despite all those failures and instabilities in detecting bounding boxes, our model can still successfully and accurately accomplish the manipulation tasks.

% \mxy{end revise}

To test our policy's robustness against this visual uncertainty, we introduce masks that randomly obscured bounding box input at varying ratios (0.0 to 1.0) during inference.

\begin{figure}[h]
\begin{center}
\includegraphics[width=0.7\linewidth]{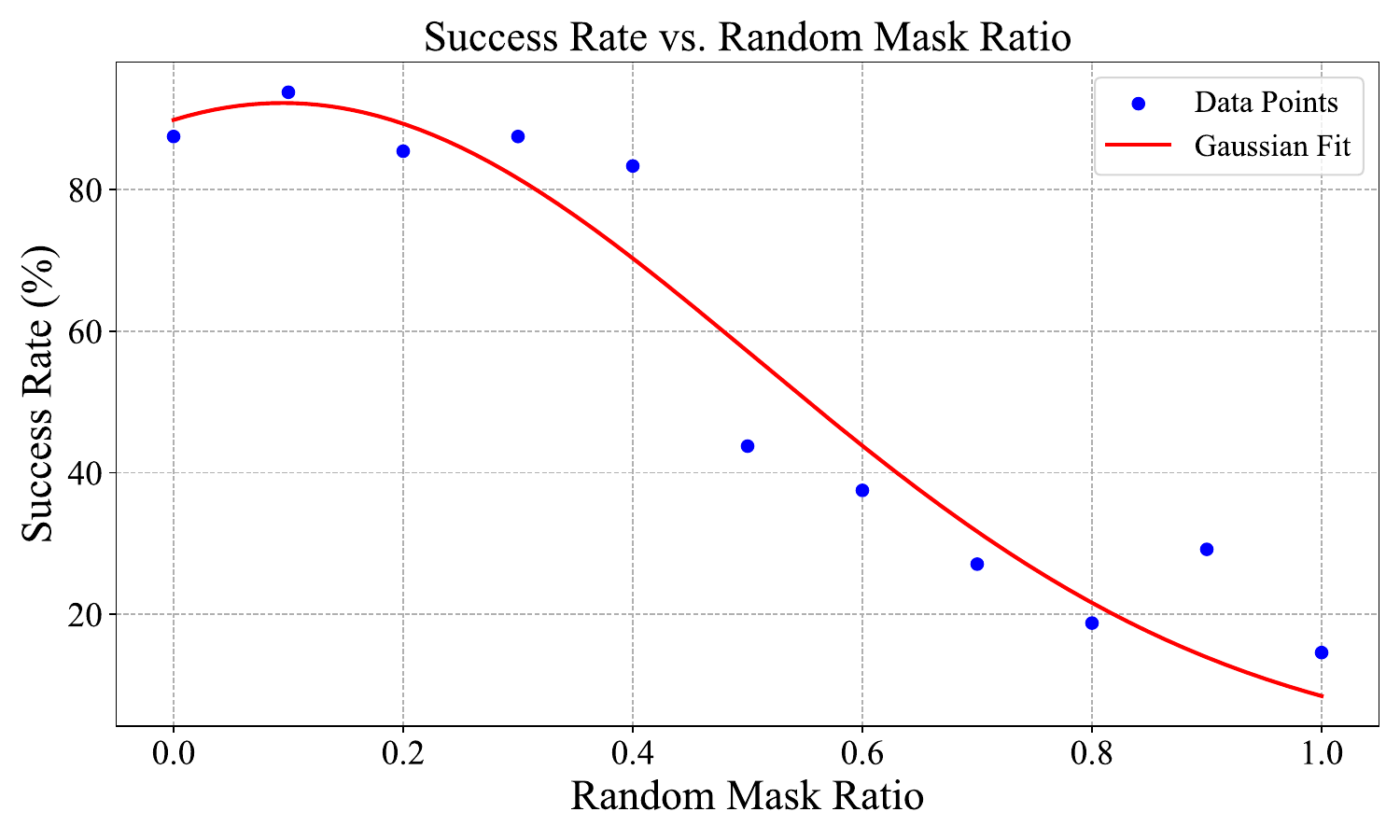}
%\framebox[4.0in]{$\; $}
% \fbox{\rule[-.5cm]{0cm}{6cm} \rule[-.5cm]{12cm}{0cm}}
\end{center}
\vspace{-0.3cm}
\caption{The relationship between random mask ratio and success rate.
The fitted curve represents a Gaussian function of $y = 92.23 \exp\left(-\frac{(x - 0.09)^2}{2 \cdot 0.41^2}\right)$.}
\label{fig:success_vs_mask_ratio}
\end{figure}

As shown in Fig.~\ref{fig:success_vs_mask_ratio}, the results reveal that the policy maintains a high success rate and generalization capability when the mask ratio is below 0.4. As the mask ratio exceeds 0.4, the success rate significantly declines, leading to consistent policy failures. 
Coincidentally, the random mask ratio we add in the student policy training phase is 0.3.
% This trend underscores the threshold at which visual uncertainty impacts the policy's performance.
This trend reveals that our approach focusing on policy generalizability can resist visual uncertainty up to 40\% of the masking ratio. That is, even if 40\% of the visual information is lost, the policy still has good spatial generalizability.
We fitted the data points with a Gaussian curve to represent this trend, defined by the equation:

\begin{equation}
\label{eq}
y(x) = a \cdot e^{-\frac{(x - b)^2}{2c^2}}
\end{equation}

where $a$, $b$, and $c$ are parameters that define the height, center, and width of the curve, respectively. This curve demonstrates that the policy is resilient up to moderate levels of visual masking (mask ratios up to 0.3), but performance degrades significantly as the masking ratio increases beyond 0.4.

This experiment validates that our policy can withstand moderate visual uncertainties during inference, maintaining robust performance.
This shows that our focus on policy generalization can easily cope with situations where the visual model has large uncertainty, leading to more robust manipulation.
However, the steep decline in success rate for higher mask ratios highlights the importance of reliable visual input for effective spatial reasoning and task execution.

As to whether the random mask in the training phase gives some increase in the success rate of our policy, please refer to  Sec.~\ref{sec:random_mask_train}.

\subsection{Random Mask during Student Policy Training Affects Performance.}
\label{sec:random_mask_train}
The uncertainty of visual detection in the real world is mentioned in the previous subsection, in order to demonstrate that random masking during training does favor the performance of the student policy, Specifically, we compare the success rates of policies trained with and without a random mask ratio of 0.3 across five different spatial ranges. During training, we randomly mask out 0 to 30\% of the bounding box information in each trajectory. This encourages the policy to adapt to scenarios where bounding box detection is incomplete or inaccurate.
Table \ref{tab:mask_comparison} demonstrated that in all range of spaces, the policy trained with random masks shows outstanding performance compared with one without masks. In addition, the policy shows better robustness in larger spatial ranges, where the variability and uncertainty of detection are more pronounced, highlighting the effectiveness of random masking in improving spatial adaptability and generalization.

\begin{table*}[h]
\centering % Use \centering for simpler alignment
\caption{Comparison of Success Rates With and Without Random Mask}
\begin{small}
\begin{tabular}{lcc}
\toprule
\textbf{Spatial Range} & \textbf{Random Mask Ratio = 0.3} & \textbf{Without Random Mask} \\
\midrule
Fix Point & \textbf{100.0\%} & \textbf{100.0\%} \\
$5~\mathrm{cm} \times 5~\mathrm{cm} \times 5~\mathrm{cm}$ & \textbf{83.33\%} & 79.17\% \\
$10~\mathrm{cm} \times 10~\mathrm{cm} \times 10~\mathrm{cm}$ & \textbf{97.92\%} & 64.58\% \\
$20~\mathrm{cm} \times 20~\mathrm{cm} \times 20~\mathrm{cm}$ & \textbf{87.5\%} & 70.83\% \\
Full Space ($41~\mathrm{cm} \times 30~\mathrm{cm} \times 28~\mathrm{cm}$) & \textbf{87.5\%} & 68.75\% \\
\bottomrule
\end{tabular}
\end{small}
\label{tab:mask_comparison}
\end{table*}

% \xxz{TODO}

\subsection{Random Noise in Bounding Boxes during Inference.} 
\label{sec:random_noise}
\begin{figure}[h]
\begin{center}
\includegraphics[width=0.7\linewidth]{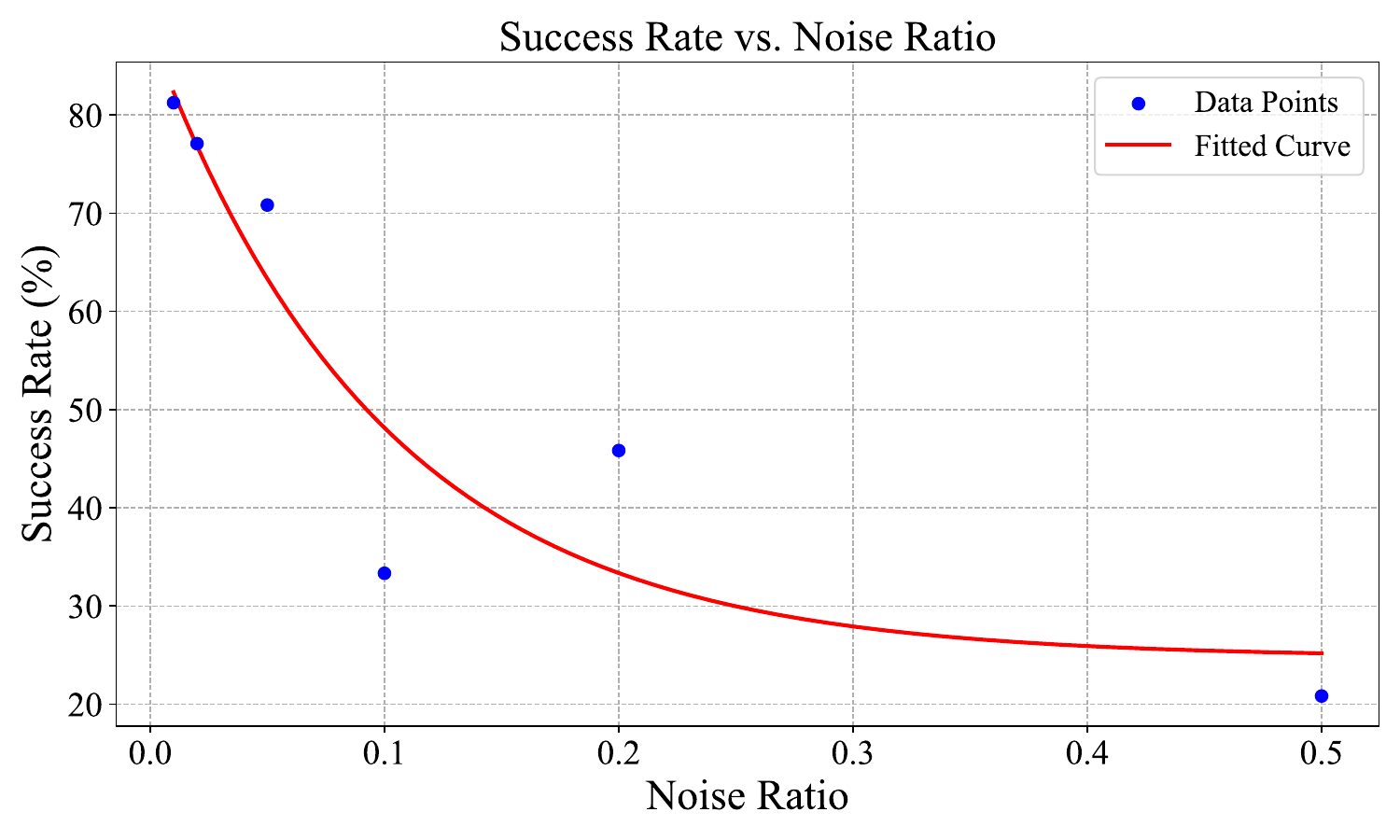}
%\framebox[4.0in]{$\; $}
% \fbox{\rule[-.5cm]{0cm}{6cm} \rule[-.5cm]{12cm}{0cm}}
\end{center}
\vspace{-0.3cm}
\caption{The relationship between noise ratio and success rate.
The fitted curve represents a exponential function of $y = 63.62 \exp(10.01 x) + 24.75$.}
\label{fig:success_vs_noise_ratio}
\end{figure}

To further examine the policy's robustness against visual uncertainty, we introduce random noise to the bounding boxes during inference. We tested noise ratios at increments of 0.01, 0.02, 0.05, 0.1, 0.2, and 0.5. Fig.~\ref{fig:success_vs_noise_ratio} illustrates the success rate as a function of noise ratio, where a monotonic decay fit highlights the policy's performance trend. The results show that the policy maintains a high success rate at low noise ratios but experienced a significant decrease as noise levels increased. 
Since our normalized bounding box value ranges from 0 to 1, the policy still achieves a high success rate with a visual detection error of 0.05, demonstrating the robustness of the policy generalization to cope with visual uncertainty and visual Sim2Real gap.

Our experiments highlight that achieving spatial intelligence requires both continuous temporal information and sufficient spatial data. 
The experiments in our main text demonstrate that spatial generalizability requires sufficient spatially varying data.
The first ablation experiment shows that continuous temporal context is crucial for effective policy performance, 
while the second and third ablation experiments demonstrate that policy generalization can cope with notable visual uncertainty.

\subsection{Comparison with IK and Other Baselines in the Simulator}
\begin{figure}[h]
\begin{center}
\includegraphics[width=\linewidth]{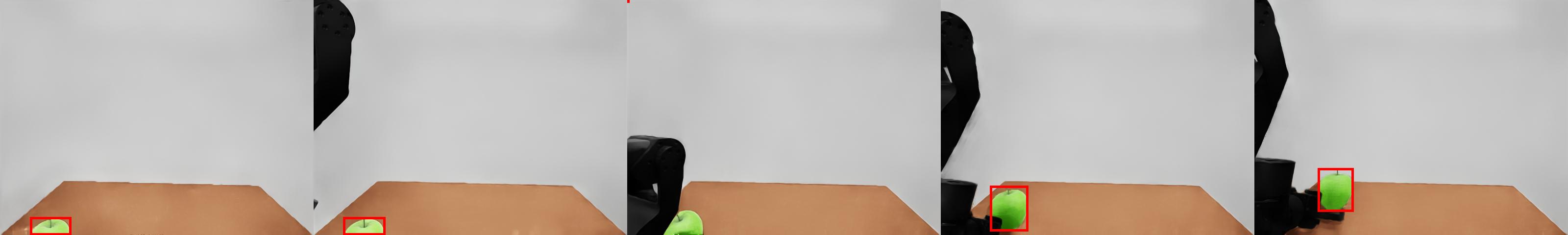}
%\framebox[4.0in]{$\; $}
% \fbox{\rule[-.5cm]{0cm}{6cm} \rule[-.5cm]{12cm}{0cm}}
\end{center}
\vspace{-0.3cm}
\caption{An example of IK failure (view of the center camera): the object is at the edge of the robotic arm's workspace - a case where the object height is too low.}
\label{pic:ik_failure_high}
\end{figure}

\begin{figure}[h]
\begin{center}
\includegraphics[width=\linewidth]{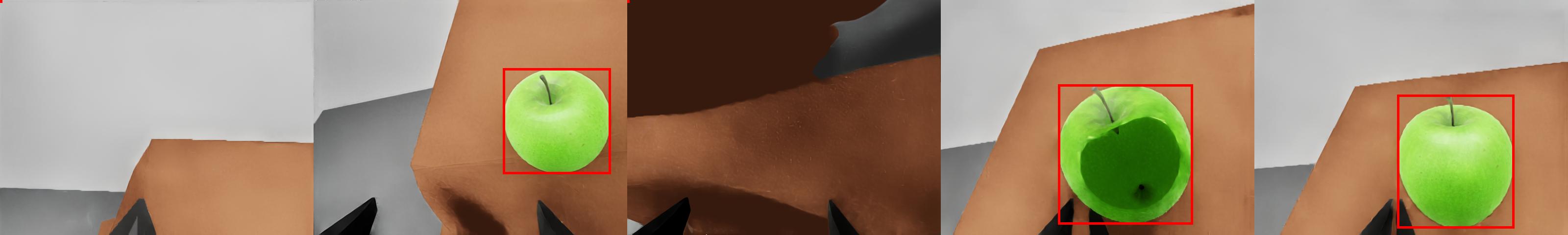}
%\framebox[4.0in]{$\; $}
% \fbox{\rule[-.5cm]{0cm}{6cm} \rule[-.5cm]{12cm}{0cm}}
\end{center}
\vspace{-0.3cm}
\caption{An example of IK failure (view of the left camera): the object is at the edge of the robotic arm's workspace - a case where the object height is too low.}
\label{pic:ik_failure_left}
\end{figure}

\begin{figure}[h]
\begin{center}
\includegraphics[width=\linewidth]{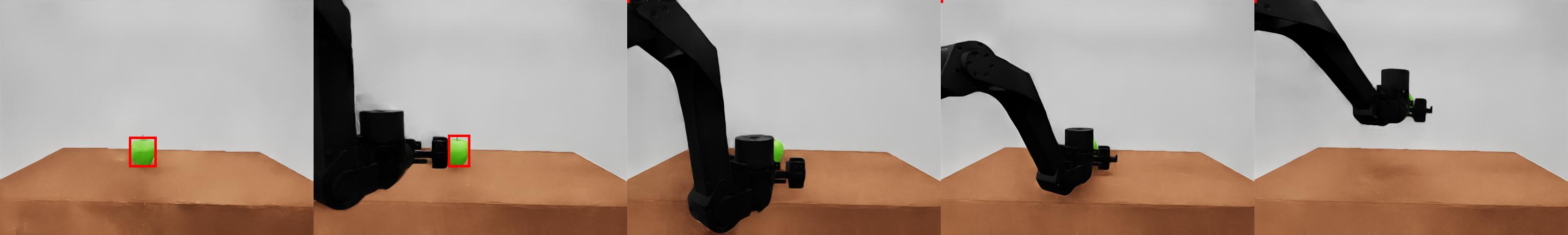}
%\framebox[4.0in]{$\; $}
% \fbox{\rule[-.5cm]{0cm}{6cm} \rule[-.5cm]{12cm}{0cm}}
\end{center}
\vspace{-0.3cm}
\caption{An example of an IK success.}
\label{pic:ik_success_high}
\end{figure}

In order to demonstrate the robustness of our method compared to other baseline methods on the grasping task, we compare our method to a heuristics-based approach, in particular the Inverse Kinematics (IK) algorithm from Isaac Lab which is required in some heuristics-based approaches such as AnyGrasp~\citep{fang2023anygrasp}.
We conduct tests using 3 different seeds in the simulator, and the comparison results are as follows:
\begin{itemize}
    \item \textbf{Inverse Kinematics (IK)}: This approach directly uses the ground truth object pose as the target end-effector (eef) pose, then computes the joint positions via IK to move the robot arm to the target. However, we observe that IK performs poorly at the edges of the workspace, especially when objects are near the table surface or at low positions. Examples of IK failures when the table is very low can be seen in Fig.~\ref{pic:ik_failure_high} and Fig.~\ref{pic:ik_failure_left}. An example of an IK success can be seen in Fig.~\ref{pic:ik_success_high}. The reason for IK failures is possible:
    \begin{itemize}
        \item The robotic arm may approach singular configurations where the Jacobian matrix becomes singular, causing unstable or undefined IK solutions.
        \item Near the workspace edge, numerical inaccuracies may result in imprecise IK solutions. Moreover, approximation methods (e.g., pseudoinverse) may struggle to converge to a valid solution in complex boundary cases.
    \end{itemize}
    \item \textbf{Key Steps}: In this approach, only key steps of the trajectory are used to model the grasping task. 
    \item \textbf{Without Random Mask}: This variant trains the student policy without random masking for object positioning.
\end{itemize}

Here, we get some key insights:
\begin{itemize}
    \item \textbf{IK Limitations}: As expected, IK struggles in the workspace boundaries due to singularities, which cause failures when the object is positioned at the edges of the table or at low heights. This limitation is inherent to traditional geometric methods like IK, which rely on precise pose knowledge and are prone to errors in challenging environments.
    \item \textbf{Our RL-Based Approach}: In contrast, \textbf{ManiBox (Ours)} utilizes RL to learn robust grasping teacher policies by directly planning the joint positions, enabling successful manipulation across a wide range of object positions. Our method does not rely on explicit pose knowledge and is highly effective even at challenging object locations that cause IK to fail.
\end{itemize}

\section{More Real-World Experiments}
\label{sec:more_exp}
\begin{figure}[h]
\begin{center}
\includegraphics[width=\linewidth]{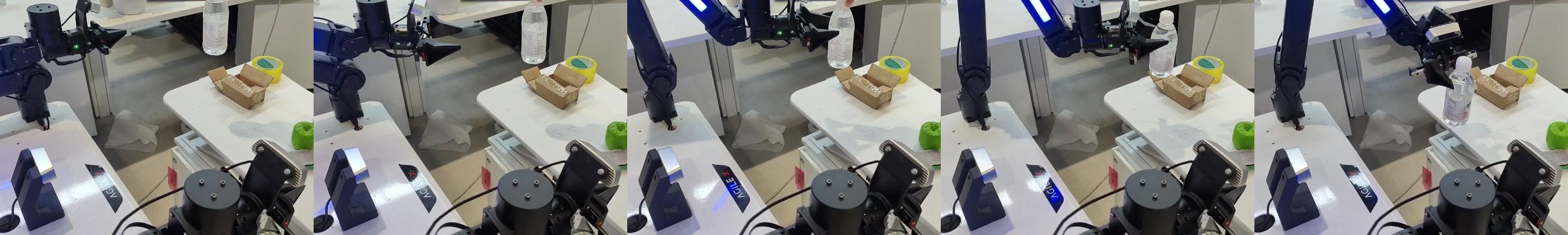}
%\framebox[4.0in]{$\; $}
% \fbox{\rule[-.5cm]{0cm}{6cm} \rule[-.5cm]{12cm}{0cm}}
\end{center}
\vspace{-0.3cm}
\caption{\textbf{Task: Mid-Air.} Grasp the bottle lifted by hand and suspended in mid-air, then move it to the target position.}
\label{mid_air}
\end{figure}

\begin{figure}[h]
\begin{center}
\includegraphics[width=\linewidth]{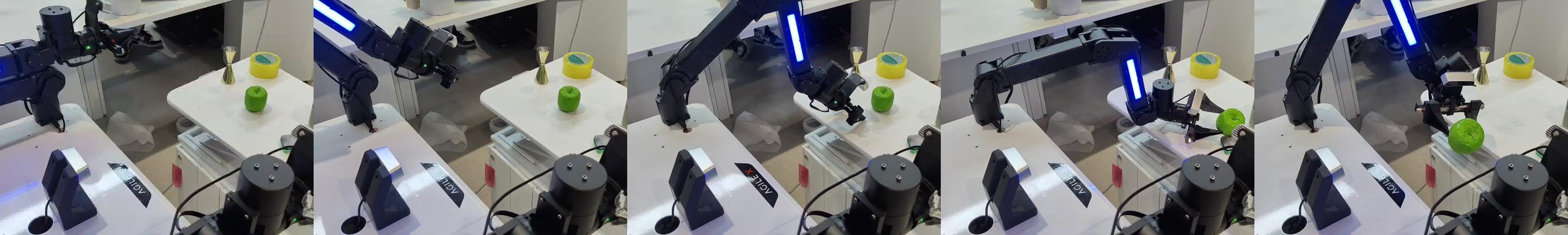}
%\framebox[4.0in]{$\; $}
% \fbox{\rule[-.5cm]{0cm}{6cm} \rule[-.5cm]{12cm}{0cm}}
\end{center}
\vspace{-0.3cm}
\caption{\textbf{Task: Cluttered Table.} Grasp the apple from a cluttered table with a height of 60 cm.}
\label{clutter_table_60}
\end{figure}

\begin{figure}[h]
\begin{center}
\includegraphics[width=\linewidth]{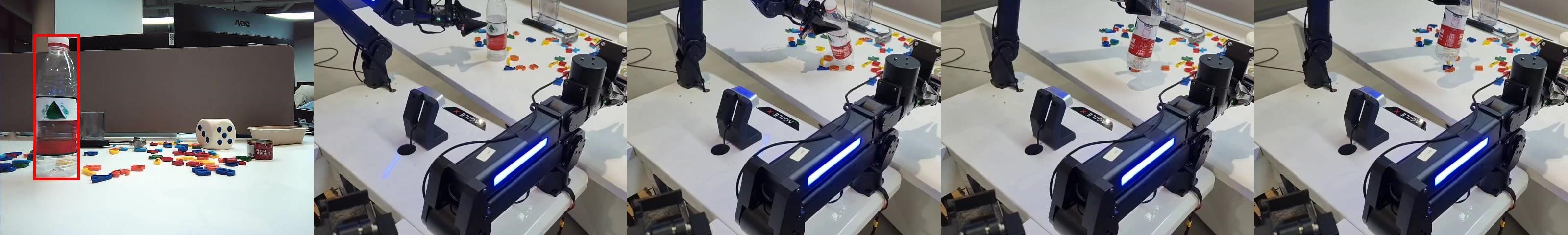}
%\framebox[4.0in]{$\; $}
% \fbox{\rule[-.5cm]{0cm}{6cm} \rule[-.5cm]{12cm}{0cm}}
\end{center}
\vspace{-0.3cm}
\caption{\textbf{Task: Cluttered Table.} Grasp the apple from a cluttered table with a height of 74.5 cm.}
\label{clutter_table_74.5}
\end{figure}

To demonstrate the extensibility of our approach, we supplement more real-world experiments:  \textbf{Mid-Air} (Figure~\ref{mid_air}) and \textbf{Cluttered Table} at varying heights (Figure~\ref{clutter_table_60} and Figure~\ref{clutter_table_74.5}).

\section{More Simulation Experiments}
\label{sec:pour_scaling_law}
\begin{figure}[t]
\begin{center}
\includegraphics[width=0.9\linewidth]{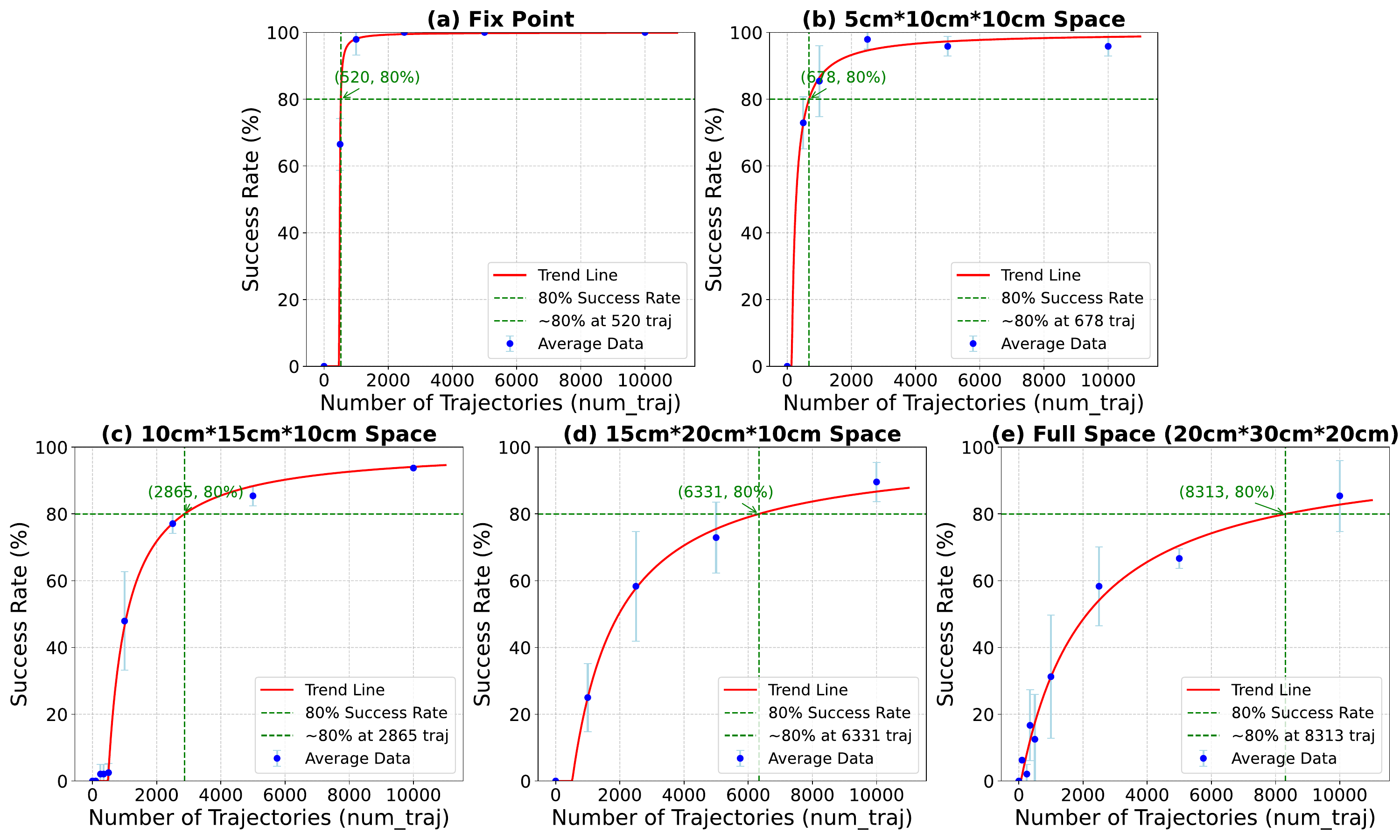}
%\framebox[4.0in]{$\; $}
% \fbox{\rule[-.5cm]{0cm}{6cm} \rule[-.5cm]{12cm}{0cm}}
\end{center}
\vspace{-0.5cm}
\caption{The scaling relationship between spatial generalization and data volume for the task of pouring water, measured under different spatial ranges. Data volume represents the number of trajectories used to train the student policy. The estimated 80\% success point is marked, with blue points showing the average success rate across 3 seeds and error bars indicating standard deviation.}
\label{fig:success_vs_data_pour}
\end{figure}

\begin{figure}[t]
\begin{center}
\includegraphics[width=0.7\linewidth]{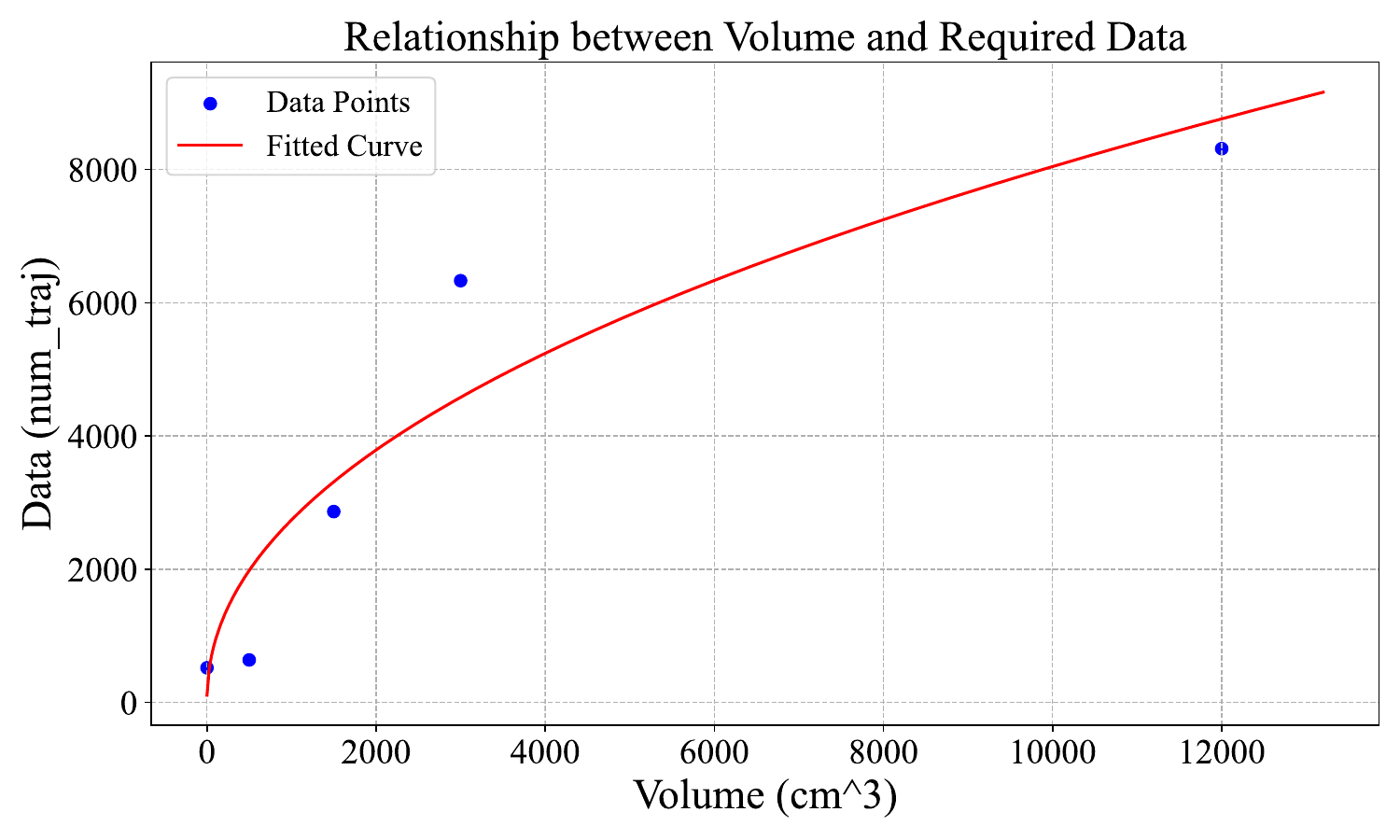}
\end{center}
\caption{The relationship between spatial volume and data amounts needed to reach 80\% grasping success rate.
The fitted curve represents a power function $y=108.24\cdot x^{0.48} $.}
\vspace{-0.5cm}
\label{fig:data_vs_volume_pour}
\end{figure}

% \thk{TODO, description}
In addition to single-object grasping, we also examine the spatial scaling law for the dual-object pouring task.
In this context, spatial volume is defined as the spatial range within which each object is randomly positioned. However, to prevent collisions, the spatial ranges of the two objects are designed to be non-overlapping.
Due to the involvement of two objects and their spatial interactions and different spatial position ranges in the pouring task, potential spatial conflicts may arise between the randomly positioned objects, leading to discrepancies between the scaling curves of pouring and grasping. 
To address this, we employ a right-shifted Michaelis-Menten curve to optimally fit the scaling relationship between spatial generalization and data volume, as shown in Fig.~\ref{fig:success_vs_data_pour}. 
Given the aforementioned task differences and the presence of noise in the experimental data, the power-law relationship between spatial volume and data amounts needed to reach 80\% grasping success rate is presented in Fig.~\ref{fig:data_vs_volume_pour}.

\section{Experimental Details}
We test the success rate of student policy in both simulator and real world.

\subsection{Evaluation in Simulation}

We run large-scale experiments in the simulator to test the spatial generalization curve for each range and determine the minimum data amounts to achieve an optimal success rate. 
To align with real robot inference, we set the robot's joints directly to target qpos instead of using environment stepping.
We mark the trajectory as success if there is a step both reaching reward $> 0.6$ and quat reward $> 1.4$.
For each policy, we test 48 trajectories and record the final success rate.
The experiment consists of three rounds of tests with different seeds, each with 16 trajectories. 
In each round, the target random points are sampled from a uniform distribution.

% \begin{figure}[h] 
% \begin{center} 
% \includegraphics[width=0.6\linewidth]{figs/new_success_rate_vs_num_traj.pdf} \end{center} \caption{The scaling relationship between spatial generalization and data amounts under different spatial ranges in the grasping tasks. The data volume represents the number of trajectories used for training the student policy. The blue points in the figure indicate the averaged success rate across three random seeds 0, 233, 16, with error bars representing the standard deviation for each point.} 
% \label{fig:success_vs_data_seed} 
% \end{figure}

% \xxz{experiments setting detail}

\subsection{Experiment Result Analysis}
We discuss our spatial scaling law here
\begin{equation}
% \text{data\_volume} = 1507 \cdot \text{spatial\_volume}^{0.28}
\text{data\_volume} = 640.32 \cdot \text{spatial\_volume}^{0.35}
\end{equation}

\begin{figure}[h]
\begin{center}
%\framebox[4.0in]{$\; $}
% \fbox{\rule[-.5cm]{0cm}{4cm} \rule[-.5cm]{12cm}{0cm}}
\includegraphics[width=0.6\linewidth]{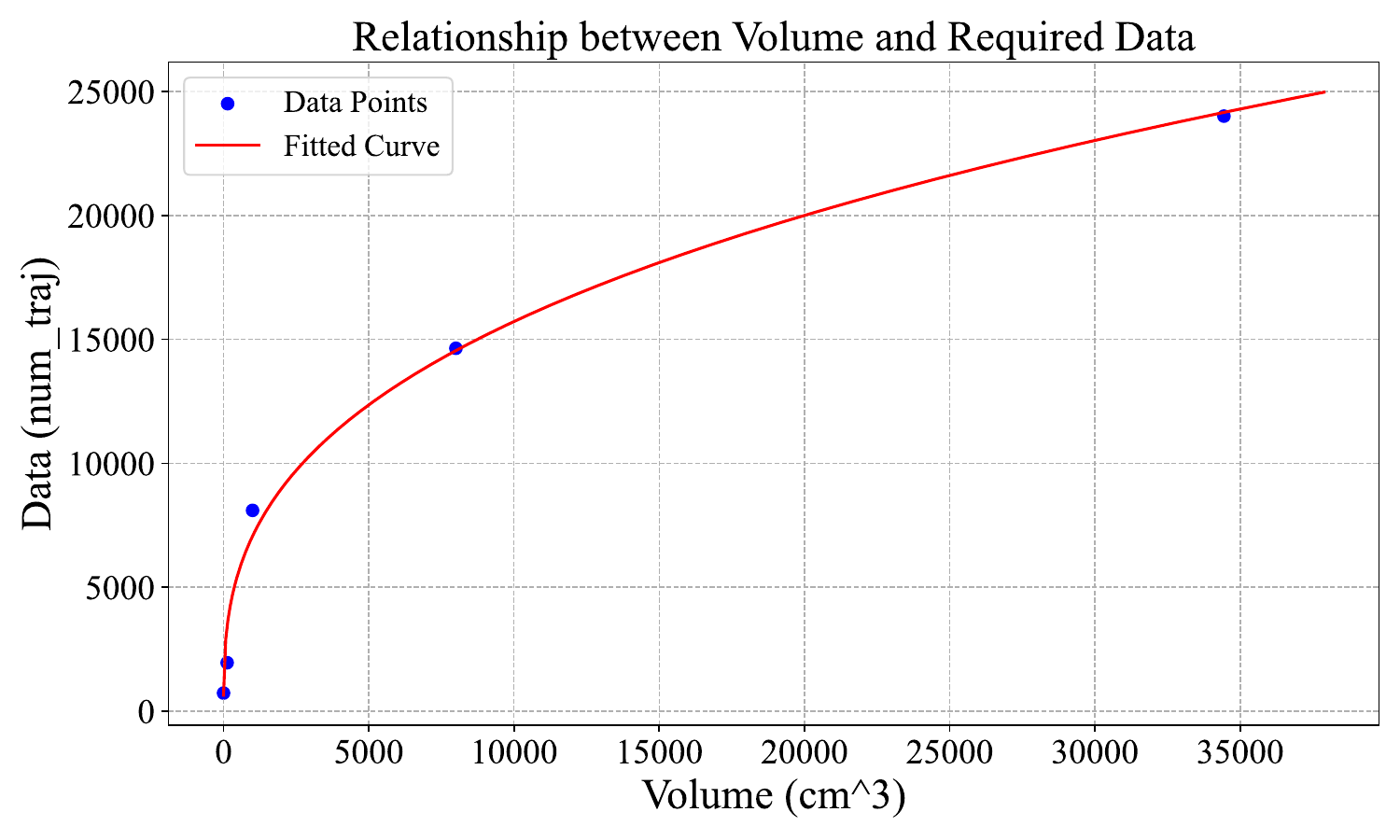}
\end{center}
\caption{Demonstration of the power-rate relationship between spatial volume and data amounts needed to reach 80\% success rate in the grasping task.}
\label{fig:app-powerlaw}
\end{figure}

\begin{figure}[h]
\begin{center}
%\framebox[4.0in]{$\; $}
% \fbox{\rule[-.5cm]{0cm}{4cm} \rule[-.5cm]{12cm}{0cm}}
\includegraphics[width=0.6\linewidth]{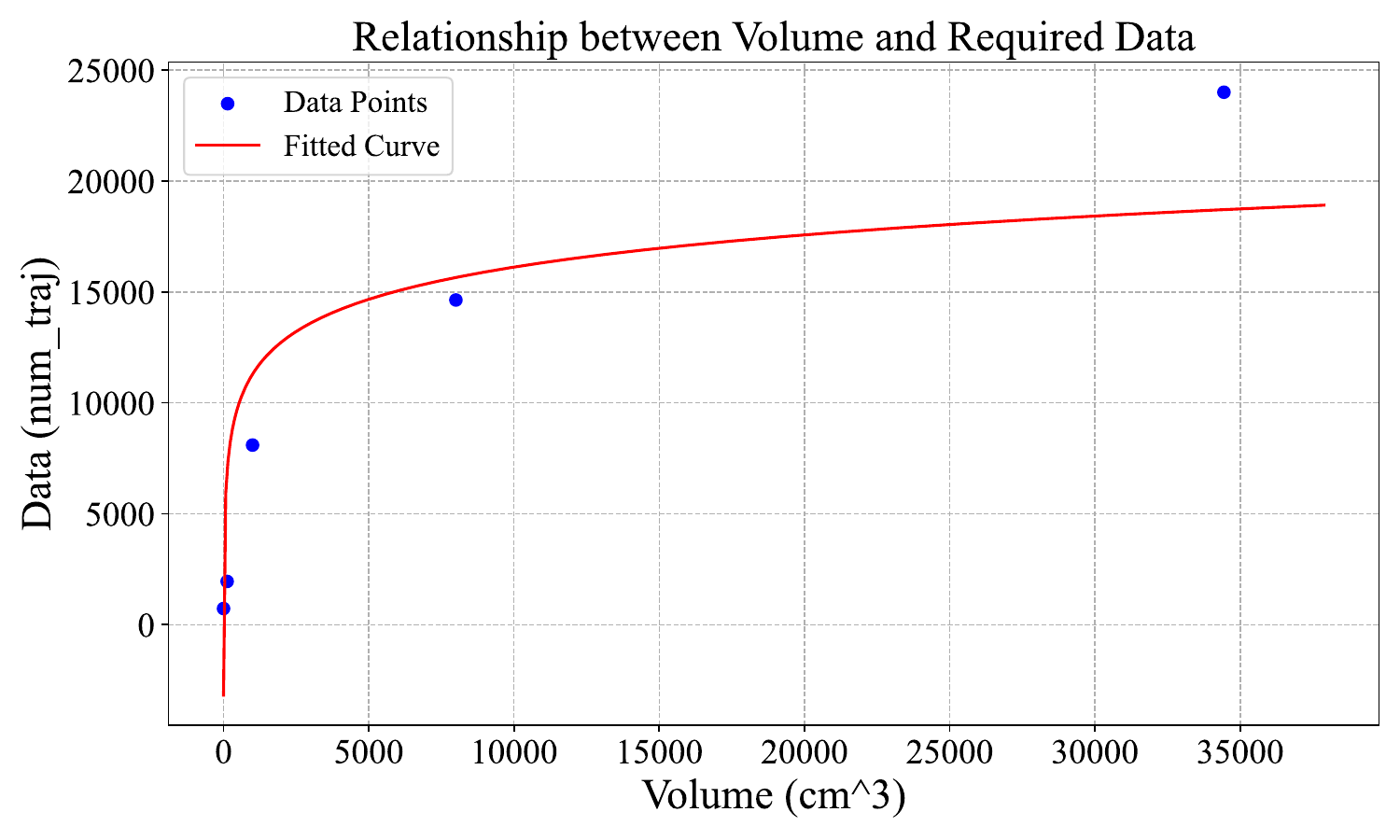}
\end{center}
\caption{Demonstration of the logarithmic relationship between spatial volume and data amounts needed to reach 80\% success rate in the grasping task.}
\label{fig:app-loglaw}
\end{figure}

By fitting the resulting data points, we observe that as the training data increases, the success rate of the policies quickly rises and eventually converges to an upper bound, which aligns well with the form of the Michaelis-Menten~\citep{michaelis1913kinetik} equation:
\begin{equation}
\text{success\_rate} = \frac{V_{\text{max}} \times \text{num\_traj}}{K_m + \text{num\_traj}}
\end{equation}
where $V_{\text{max}}$ represents the maximum achievable success rate, and $K_m$ is the number of trajectories at which the success rate is half of $V_{\text{max}}$. This equation captures the saturation-type growth observed in the policy performance.

% Across different settings, the amount of data required to achieve an 80\% success rate shows a power function as the data increases:

% $y=1507\cdot x^{0.28} $.
% % The fitted curve represents a logarithmic function of \textit{a\cdot \ln(b \cdot x) + c}.

For the functional relationship between the amount of data required for spatial generalization and spatial volume, we explored two types of fits: a power law and a logarithmic function. 
The fitting results of power law is given by $y = a\cdot x^b= 640.32 x^{0.35}$
% $y = a\cdot x^b= 1507 x^{0.28}$
, while the logarithmic fit is 
% $y = a\cdot \ln(b \cdot x) + c = 2450.69 \cdot \ln{(0.51 \cdot x)}  - 1604.28$. 
$y = a\cdot \ln(b \cdot x) + c = 2095.51 \cdot \ln{(0.3 \cdot x)}  - 642.93$. 
We found that the logarithmic fit becomes unreasonable as the spatial volume $x$ approaches 0, where the required data, 
% $y = 2450.69 \cdot \ln{0.51 \cdot x}  - 1604.28$
$y = a\cdot \ln(b \cdot x) + c = 2095.51 \cdot \ln{(0.3 \cdot x)}  - 642.93$
, tends to negative infinity, which is unrealistic since the amount of data cannot be negative. 
Furthermore, the fit results in a negative value at $x = 3.3$,
% $x = 2$, 
where $bx = 1$ and $\log(bx) = 0$, leading to a required data amount of 
% $a \log(bx) + c = -1604$. 
$a \log(bx) + c = -642.93$. 
In the long run (for larger spatial volumes), the power law fit better describes the relationship between the data required for spatial generalization and spatial volume. Fig.~\ref{fig:app-powerlaw} shows the selected power law relationship, and Fig.~\ref{fig:app-loglaw} displays the logarithmic curve.

\subsection{Evaluation in Real World}
\label{app:real_eval}

\subsubsection{Real-World Success Rate}
To verify our experiment results from the simulator, we evaluate our policy in the real robot. 
For each range, we choose the first optimal policy that achieves 80\% success rate in simulator and evaluate with the real robot for 10 randomly chosen points.

\paragraph{Random Placement of Objects in Real-world Evaluation.} To be more specific, we divide the maximum range, namely the full space ( $41cm*30cm*28cm$), into three equal parts in height, two equal parts in width, and three equal parts in length, resulting in a total of $3*2*3=18$ equally sized spatial cuboid ranges at different positions.
For each tested spatial range, ten different points are uniformly selected to test the robot's manipulation success rate.

\subsubsection{Real-World Experiment Settings}
% To verify our methods, we evaluate our policy in the real robot. 
% For each range, we choose the first optimal policy that achieves 80\% success rate in simulator and 
To verify our methods, we evaluate the student policy in the real robot, which is trained on simulation data of full space ( $41\mathrm{cm}*30\mathrm{cm}*28\mathrm{cm}$) with the real robot.

% \mxy{TODO, spatial generalization, We are randomly selecting points in each region C, LB...}
\paragraph{Spatial Generalization.} As illustrated in Fig.~\ref{fig:demo_fig}, we divide the spatial range horizontally into Left (L), Right (R), and Center (C), and vertically into Front (F) and Back (B), and randomly select points within these sections on different heights to test our policy\footnote{The table is an elevated table and can therefore be adjusted to any height.}.
All these points are arbitrarily selected, which means that the target object has an equal chance of appearing at the center or the edge of the table, far from or close to the robot, in positions that require the robotic arm to reach upwards or downwards.
We hope this approach reflects the spatial generalization of our policy comprehensively and honestly.

% \mxy{end revise}

In addition to the verification of spatial generalization, we also test the ability of Object generalization and background generalization. The details of the experiment setting can be found in Tab.~\ref{tab:experiment_settings}.

% \revise{object generalization}

% \mxy{TODO, Explain that the domain randomization we do can generalize to objects of unseen size.(object generalization)}
\paragraph{Object Generalization and Background Generalization.}Furthermore, we have chosen objects of various shapes, sizes, and materials as target objects for testing, all of which were previously unseen by our policy. 
To be more specific, during training, we only encountered apples with sizes of approximately $(6.9cm-7.4cm, 6.9cm-7.4cm, 6.9cm-7.4cm)$ in the XYZ direction, while the test target objects span a size range of $(5.1cm-7.0cm, 5.1cm-9.7cm, 6.0cm-22.7cm)$ in the XYZ direction.
Additionally, although apples are roughly spherical, the target objects also include cylindrical and pyriform shapes.
Moreover, the target objects encompass different materials such as glass and metal.
All these objects serve to demonstrate our policy's capability to generalize to varying shapes, sizes, and materials.

% \mxy{end revise}

\begin{table*}[h]
\caption{Real-World Experiment Settings. Note that we use only the bounding boxes of the objects for both student policy training and inference. Bolded object sizes are never seen in training.}
\label{tab:experiment_settings}
% \vskip 0.15in
\begin{center}
\begin{small}
\begin{tabular}{lll}
\toprule
\textbf{Generalization Type} & \textbf{Object Size (cm) of XYZ}  & \textbf{Setting Description} \\
\midrule
\textbf{Object in Simulation Data} & \\
Apple (Domain Randomization) & (6.9-7.4, 6.9-7.4, 6.9-7.4) & Distribute uniformly in a configurable spatial volume \\
\midrule
\textbf{Object Generalization} & \\
Yellow Pear & \textbf{(6.5, 5.5, 10.7)} & h4: Point Random, Minimal setting \\
Orange & \textbf{(7.0, 6.7, 6.5)} & h4: Point Random, Minimal setting \\
Steel Cup & \textbf{(6.5, 9.7, 10.8)} & h4: Point Random, Minimal setting \\
Tin Can & \textbf{(5.1,5.1,7.2)}  & h4: Point Random, Minimal setting \\
Glass Beaker & \textbf{(5.1, 5.1, 6.0)} & h4: Point Random, Minimal setting \\
\midrule
\textbf{Background Generalization} & \\
Apple & (7.2, 7.4, 6.9) &  h4: Point Random, Orange tablecloth \\
Unseen Apple & \textbf{(7.0, 7.5, 7.0)} & h4: Point Random, Black tablecloth \\
Apple & (7.2, 7.4, 6.9) & h4: Point Random, Playing video \\
Apple & (7.2, 7.4, 6.9) & h4: Point Random, Dim light \\
Apple & (7.2, 7.4, 6.9) & h4: Point Random, Flittering light \\
\midrule
\textbf{Complex Generalization} & \\
Unseen Apple, Unseen Table & \textbf{(7.0, 7.5, 7.0)} & Point Random, Table 2 \\
Unseen Apple, Unseen Desk & \textbf{(7.0, 7.5, 7.0)} & Point Random, Multiple objects \\
Steel Cup, Unseen Desk & \textbf{(6.5, 9.7, 10.8)}  & Point Random, Daily desktop \\
Plastic Bottle, Unseen Desk & \textbf{(6.4, 6.4, 22.27)} & Point Random, Daily desktop \\
\bottomrule
\end{tabular}
\end{small}
\end{center}
% \vskip -0.1in
\end{table*}

The three camera images, as well as the entire trajectory of grasping, are schematized below in Figure~\ref{real_multiview}:
% \thk{more pictures of first-person view}
\begin{figure}[h]
\begin{center}
%\framebox[4.0in]{$\; $}
% \fbox{\rule[-.5cm]{0cm}{4cm} \rule[-.5cm]{12cm}{0cm}}
\includegraphics[width=\linewidth]{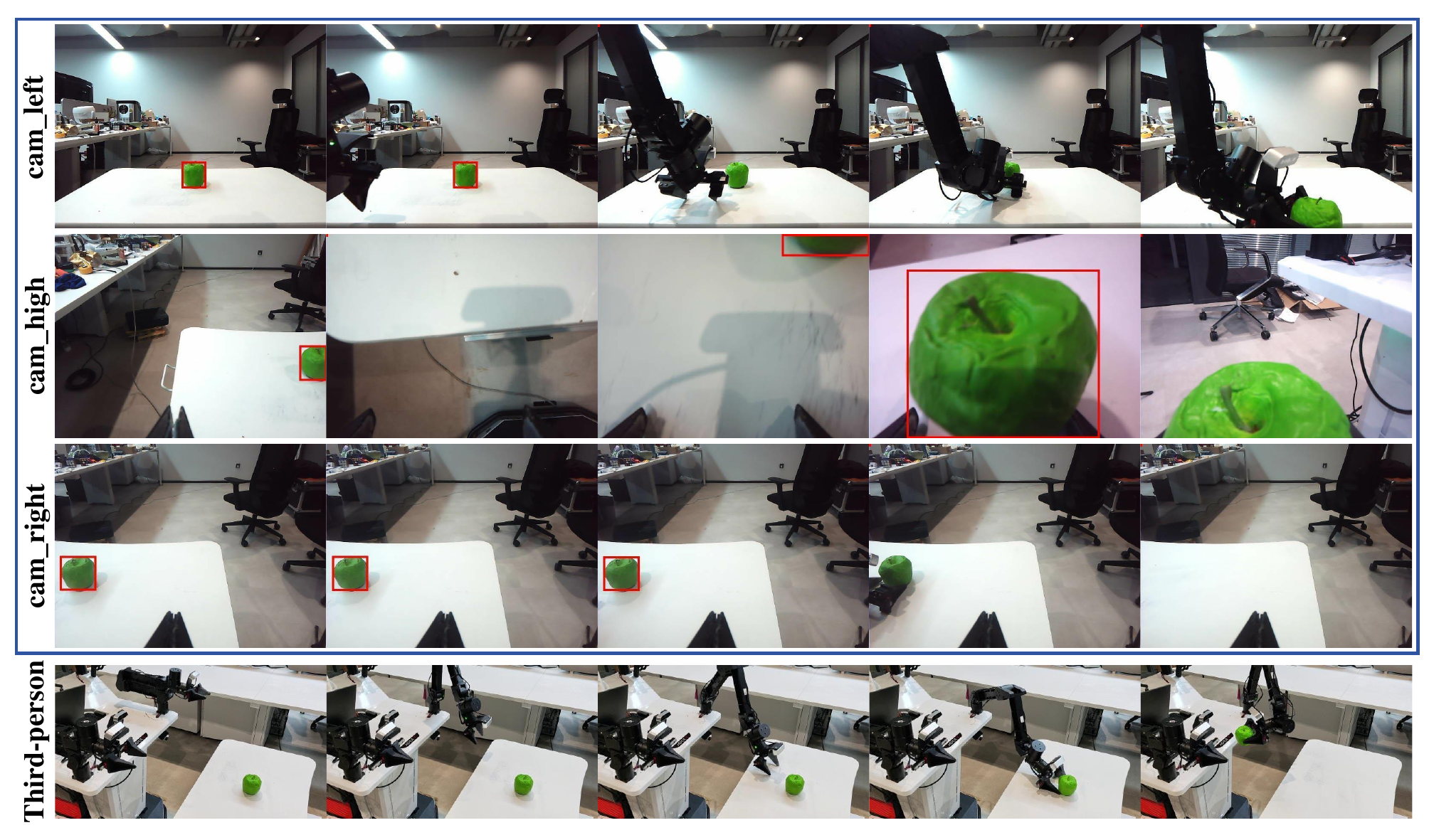}
\end{center}
\caption{Pictures of the entire trajectory of grasping, including perspective and three camera views.}
\label{real_multiview}
\end{figure}

\subsection{More Real-world Results}
\label{app_real_world_exp}

% \begin{wraptable}{r}{9.9 cm}
\begin{table}[H]
\centering
% \fontnotesize
% \vspace{-.4cm}
\caption{The real-world success rate (SR) of Manibox and vision-based ACT under varied spatial volumes. We select an approximate data amount based on the results in Fig.~\ref{fig:success_vs_data}. 
}     
% \vspace{-.1cm}
% \smallskip
\small
% \resizebox{0.70\textwidth}{!}{ 
\begin{tabular}{cll}
\toprule
\textbf{Spatial Range}      & \textbf{Method} & \textbf{SR} \\ \hline
\multirow{2}{*}{Fixed Point} & Vision-Based ACT & 70\% \\
& ManiBox (Ours) & 100\% \\ \hline
\multirow{2}{*}{5cm*5cm*5cm} & Vision-Based ACT & 0\% \\
& ManiBox (Ours) & 70\% \\ \hline
\multirow{2}{*}{10cm*10cm*10cm} & Vision-Based ACT & 0\% \\
& ManiBox (Ours) & 100\% \\ \hline
\multirow{2}{*}{20cm*20cm*20cm} & Vision-Based ACT & - \\
& ManiBox (Ours) & 80\% \\ \hline
\multirow{4}{*}{\makecell[c]{Full Space\\(41cm*30cm*28cm)}} & Vision-Based ACT & - \\
&\makecell[c]{Vision-Based ACT \\(100 real-world data) } & 0\% \\ 
& ManiBox (Ours) & 90\% \\ \bottomrule
\end{tabular}
% }
\label{real_success_table}
% \vspace{-.4cm}
\end{table}
% \end{wraptable}

% \section{Details of Simulations}
\section{Teacher-Student Training Details}
For ease of understanding, Alg.~\ref{alg:manibox} is the pseudo algorithm for the ManiBox method. 

% \begin{algorithm}[h]
%    \caption{ManiBox}
%    \label{alg:example}
% \begin{algorithmic}
   %  \STATE Initialize $env$ in simulator.
   %  \STATE Teacher Policy $\pi_\beta$ = PPO().
    
   % \STATE {\bfseries Input:} data $x_i$, size $m$
   % \REPEAT
   % \STATE Initialize $noChange = true$.
   % \FOR{$i=1$ {\bfseries to} $m-1$}
   % \IF{$x_i > x_{i+1}$}
   % \STATE Swap $x_i$ and $x_{i+1}$
   % \STATE $noChange = false$
   % \ENDIF
   % \ENDFOR
   % \UNTIL{$noChange$ is $true$}
% \end{algorithmic}
% \end{algorithm}

\begin{algorithm}
\caption{Training and Deployment Algorithm of ManiBox}\label{alg:manibox}
\begin{algorithmic}[1]
\Require Simulator environment $Sim$, Real robot $Robot$
\Require Proximal Policy Optimization (PPO) for training teacher policy
\Require Bounding box extractor $BBoxExtractor$
\Require Maximum dataset size $N_{max}$, reward function $R$, domain randomization module $DR$
\Statex

\State \textbf{Stage 1: Teacher Policy Training}
\State Initialize teacher policy $\pi_\beta$ with random weights
\State Set $Sim$ environment with reward function $R$ and domain randomization module $DR$
\While{not converged}
    \State Collect trajectory data $\mathcal{D}_T$ in $Sim$ using $\pi_\beta$
    \State Update $\pi_\beta$ using PPO with $\mathcal{D}_T$
\EndWhile

\Statex

\State \textbf{Stage 2: Simulation Data Generation}
\State Initialize dataset $\mathcal{D}_S = \emptyset$
% \For{each task configuration in $Sim$}
\While{$|\mathcal{D}_S| < N_{max}$}
    \State Reset $Sim$ environment
    \State Apply domain randomization $DR$ to environment
    \State Generate trajectory $\tau = \{(o^{rob}_t, a_t, r_t)\}_{t=1}^T$ using $\pi_\beta$
    \State Extract bounding boxes $\{o^{vis}_t\}_{t=1}^T$ using $BBoxExtractor$
    \State Store $\{(o^{vis}_t, o^{rob}_t, a_t)\}_{t=1}^T$ in $\mathcal{D}_S$
\EndWhile

\Statex

\State \textbf{Stage 3: Student Policy Training}
\State Initialize student policy $\pi_\theta$ with random weights
\While{not converged}
    \State Sample trajectory $\tau = \{(o^{vis}_t, o^{rob}_t, a_t)\}_{t=1}^T$ from $\mathcal{D}_S$
    \State Randomly mask some $o^{vis}_t$ to simulate detection failures
    \State Update $\pi_\theta(o^{vis}_{\le t}, o^{rob}_{\le t})$ using the masked $\tau$
\EndWhile

\Statex

\State \textbf{Stage 4: Real-World Deployment}
\State $t\gets 0$
\While{not terminated}
    \State Extract real-world bounding boxes $o^{vis}_t$ using $BBoxExtractor$
    \State Compute action $a_t \sim \pi_\theta(o^{vis}_{\le t}, o^{rob}_{\le t})$
    \State Execute action $a_t$ on $Robot$
    \State $t\gets t+1$
\EndWhile
\end{algorithmic}
\end{algorithm}

% \section{Teacher-Student Training Details}

% To maximize the speed of training and data collection in the simulator, we choose the highly parallelized Isaac Lab~\citep{mittal2023orbit} as the simulation environment. \mxy{Already mentioned in Teacher Policy}
% The training of our teacher policy and data generation are carried out in the simulator with high parallelism. The generated simulator data, which includes bounding boxes, is used to train the student policy.

\subsection{Teacher Policy Details}
\label{app_simulation}
% tricks of PPO (empirical normalization...), reward, ...

\begin{figure}[H]
\begin{center}
%\framebox[4.0in]{$\; $}
% \fbox{\rule[-.5cm]{0cm}{4cm} \rule[-.5cm]{12cm}{0cm}}
\includegraphics[width=\linewidth]{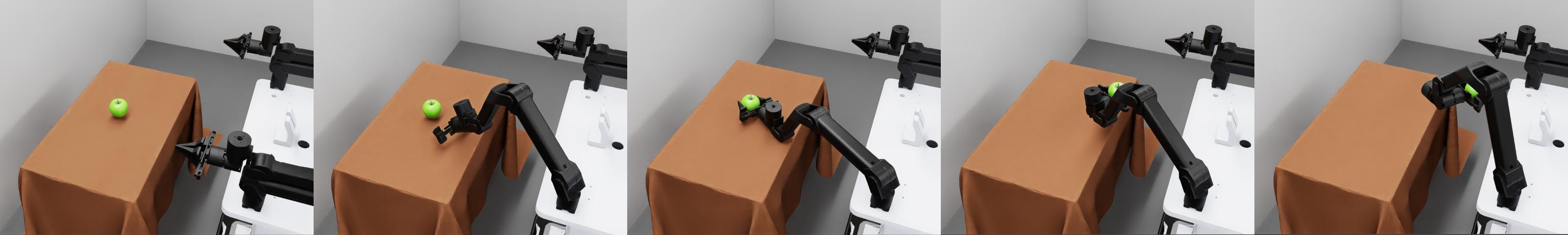}
\end{center}
\caption{The simulation environment consists of a room with a table, an apple, and a robot. The task is for the robot to grasp the apple and place it in a specified location.}
\label{perspective_view}
\end{figure}

% In this section, we will introduce more details about the simulation, including the reward function design.
We train our teacher policy using PPO with 8,192 parallel environments in Isaac Lab~\citep{mittal2023orbit}. 
With a carefully designed reward function and applied PPO techniques, the training is completed in around 40 minutes, reaching approximately 2,000 training iterations.
The task involves training a robot to grasp an apple randomly placed on a liftable table and transport it to a designated target position (Figure \ref{perspective_view}). 

The advantages of using a teacher policy instead of direct visual RL training can be summarized in three key aspects. 
First, the teacher policy relies solely on robot and object information as input, significantly reducing the dimensionality of the state space. 
Second, the incorporation of privileged information and tailored reward functions for critical steps offers more focused guidance during training, thereby narrowing the exploration space. Lastly, visual RL suffers from limited parallelizability, which results in lower overall training efficiency.
% \paragraph{Details of Simulations} 
% A more detailed description of the simulation tasks and object setups would help understand the system’s scope

\subsubsection{Simulation Scenario Setup} 
\label{sec:sim_scene_setup}
\begin{figure}[H]
\begin{center}
%\framebox[4.0in]{$\; $}
% \fbox{\rule[-.5cm]{0cm}{4cm} \rule[-.5cm]{12cm}{0cm}}
\includegraphics[width=\linewidth]{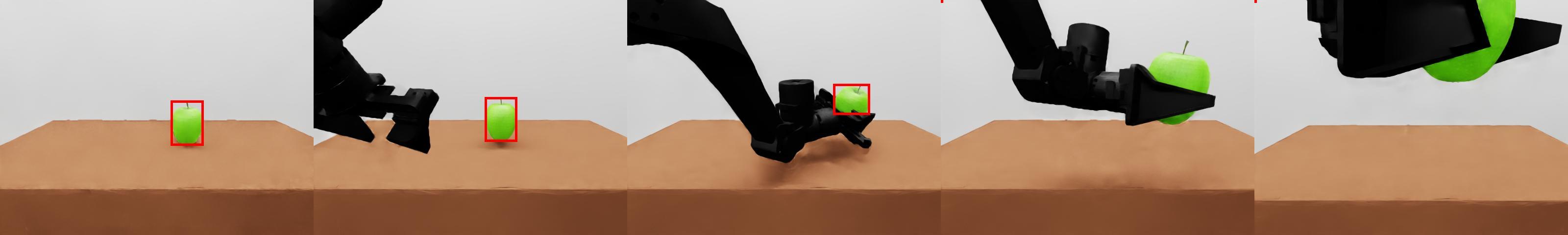}
\end{center}
\caption{Key steps of grasping in simulation in cam\_high view.}
\label{cam_high_view}
\end{figure}

\begin{figure}[H]
\begin{center}
%\framebox[4.0in]{$\; $}
% \fbox{\rule[-.5cm]{0cm}{4cm} \rule[-.5cm]{12cm}{0cm}}
\includegraphics[width=\linewidth]{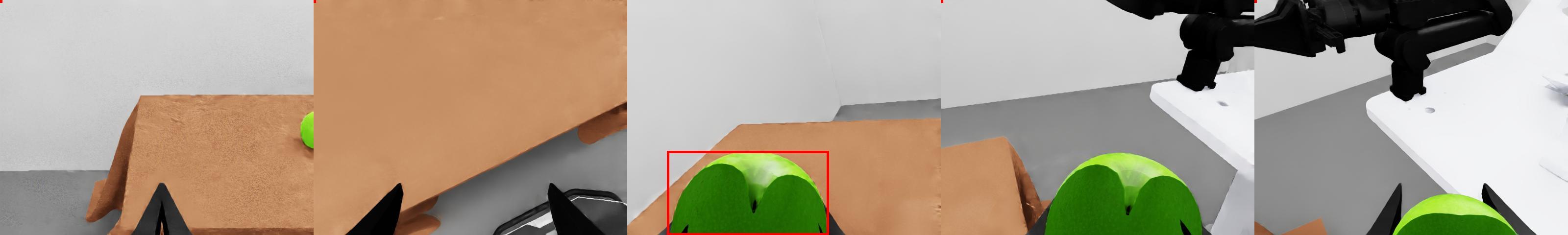}
\end{center}
\caption{Key steps of grasping in simulation in cam\_left\_wrist view.}
\label{cam_left_view}
\end{figure}

\begin{figure}[H]
\begin{center}
%\framebox[4.0in]{$\; $}
% \fbox{\rule[-.5cm]{0cm}{4cm} \rule[-.5cm]{12cm}{0cm}}
\includegraphics[width=\linewidth]{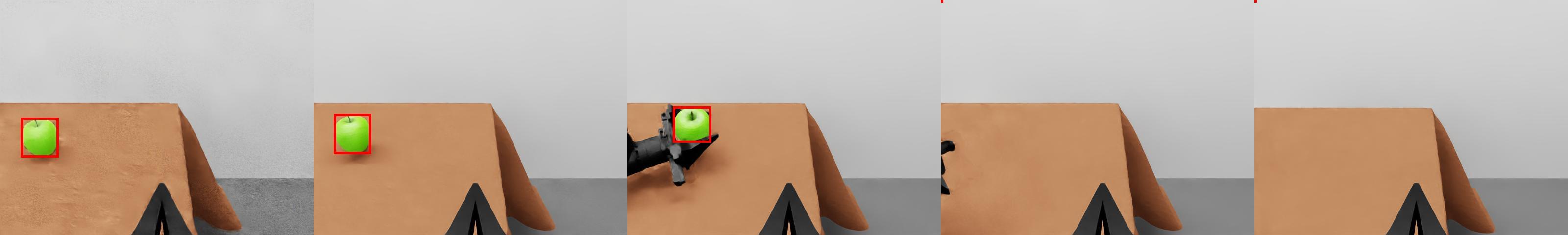}
\end{center}
\caption{Key steps of grasping in simulation in cam\_right\_wrist view.}
\label{cam_right_view}
\end{figure}

The environment is built based on the lift task in Isaac Lab.  
During training, the scene consists of a table, an apple, and a robot. 
The robot is initialized at a fixed starting position at the beginning of each episode. The apple's size, as well as its x,y position, are uniformly randomized across environments, while the height of the table is also uniformly randomized. The target position, however, remains fixed throughout all training episodes. Key steps captured from three camera perspectives are shown in Figures~\ref{cam_high_view}, \ref{cam_left_view}, and \ref{cam_right_view}.
% The apple's scale is uniformly randomized across environments, while its spatial position and the height of the table are uniformly randomized for each individual run.  
To support future extensions to more complex tasks, such as obstacle avoidance during grasping, we represent the robot's behavior with a 30Hz trajectory.  
Each environment in the simulator is evenly spaced to prevent interference between different training instances.  
Further details on the randomized ranges can be found in Table~\ref{range_table}.

\begin{table*}[h]
\vspace{-1em}
\caption{Environment Randomization Parameters}
\vspace{-1em}
\label{range_table}
% \vskip 0.15in
\begin{center}
\begin{small}
\begin{tabular}{lccc}
\toprule
\textbf{Spatial Range} & \textbf{Table Height (z)} & \textbf{Apple X Position} & \textbf{Apple Y Position} \\
\midrule
Fix Point & $(-0.00, 0.00)$ & $(-0.0, 0.0)$ & $(-0.0, 0.0)$ \\
$5 \, \text{cm} \times 5 \, \text{cm} \times 5 \, \text{cm}$ & $(-0.025, 0.025)$ & $(-0.025, 0.025)$ & $(-0.025, 0.025)$ \\
$10 \, \text{cm} \times 10 \, \text{cm} \times 10 \, \text{cm}$ & $(-0.07, 0.03)$ & $(-0.05, 0.05)$ & $(-0.05, 0.05)$ \\
$20 \, \text{cm} \times 20 \, \text{cm} \times 20 \, \text{cm}$ & $(-0.1, 0.1)$ & $(-0.1, 0.1)$ & $(-0.05, 0.15)$ \\
Full Space ($41 \, \text{cm} \times 30 \, \text{cm} \times 28 \, \text{cm}$) & $(-0.13, 0.15)$ & $(-0.22, 0.08)$ & $(-0.05, 0.36)$ \\
\midrule
\textbf{Apple 3D Scale Range} & \multicolumn{3}{c}{$((0.77,0.82), (0.77,0.82), (0.77,0.82))$} \\
\bottomrule
\end{tabular}
\end{small}
\end{center}
% \vskip -0.1in
\end{table*}

\subsubsection{Manipulation Tasks on Two Objects}
Manipulation tasks involving two objects require the detection model to recognize the two objects and then perform the corresponding skills. And a policy is trained to master a specific manipulation skill, such as pouring, grasping, or placing. For example, the pouring policy recognizes two bounding boxes, “bottle” and “cup”, and then the policy will grab the bottle and pour it into the cup. 
Through the decoupling of the teacher policy and student policy, our approach offers significant flexibility in training the teacher policy. 
This flexibility allows for the utilization of both RL methods, as previously discussed, and alternative approaches such as state machines. 
Given the inherent challenges of training RL for multi-object, long-horizon manipulation tasks, ManiBox employs a state machine as the teacher policy specifically for the pouring water task. Notably, the remaining components of the system remain unchanged, ensuring consistency and reliability in the overall framework.

\begin{figure}[H]
\begin{center}
\includegraphics[width=\linewidth]{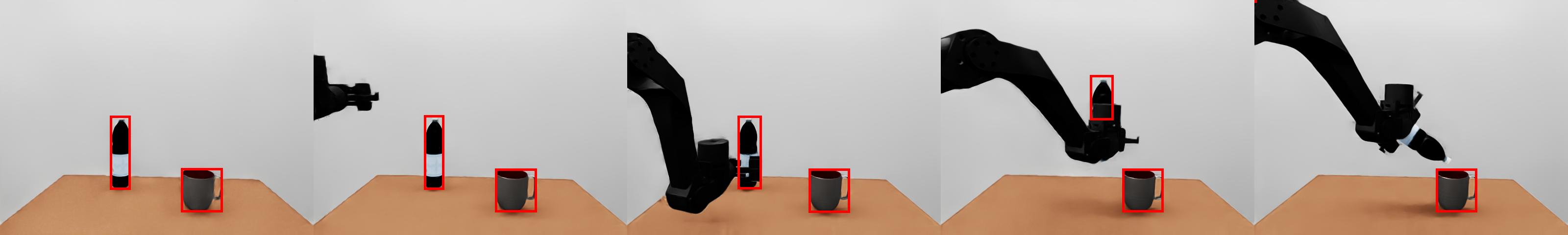}
\end{center}
\caption{Key steps of pouring water in simulation in cam\_high view.}
\label{fig:pour_water_sim_traj}
\end{figure}

In the task of pouring water, the x,y positions of the bottle and cup are uniformly randomized across environments, while the height of the table is also uniformly randomized. Key steps captured in the simulator are shown in Fig.~\ref{fig:pour_water_sim_traj}.

\begin{figure}[H]
\begin{center}
\includegraphics[width=\linewidth]{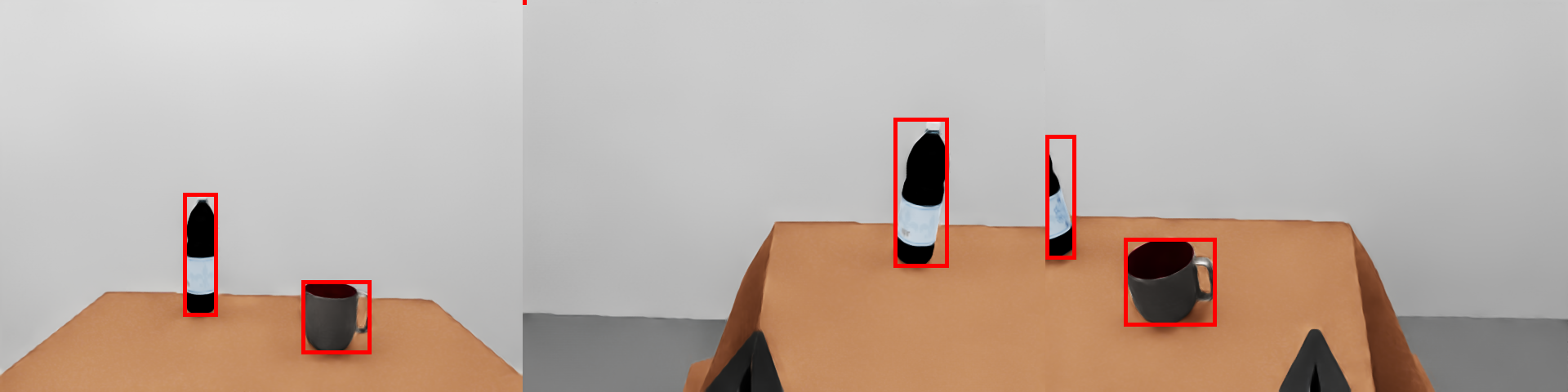}
\end{center}
\caption{Three views of the initial timestep of the pouring water task, in order: cam\_high, cam\_left\_wrist, and cam\_right\_wrist.}
\label{fig:pour_water_sim_3views}
\end{figure}

\subsubsection{Details of Observation and Action}

We provide detailed information about the observations and actions used in training the teacher policy for the multi-environment object-grasping manipulation task:

\begin{itemize}
    \item \textbf{Observations:}
    \begin{itemize}
        \item Contact forces (24-dimensional)
        \item Joint position (8-dimensional)
        \item Joint velocity (8-dimensional)
        \item Object position (3-dimensional)
        \item Target location (7-dimensional)
        \item Last action (8-dimensional)
    \end{itemize}
    \item \textbf{Actions:}
    \begin{itemize}
        \item Joint Position (8-dimensional) (We choose position control. The action is the expected joint position for the next timestep.)
    \end{itemize}
\end{itemize}

\subsubsection{Implementation of PPO}
Our PPO implementation is built on \href{https://github.com/leggedrobotics/rsl\_rl}{rsl\_rl}, and leverages several techniques to improve stability and performance. Key components and their settings are summarized in Table~\ref{tab:ppo_parameters}.

\begin{table*}[h]
\vspace{-1em}
\caption{PPO Parameters}
\vspace{-1em}
\label{tab:ppo_parameters}
% \vskip 0.15in
\begin{center}
\begin{small}
\begin{tabular}{lll}
\toprule
\textbf{Techniques} & \textbf{Parameter} & \textbf{Value} \\
\midrule
Clipped Surrogate Objective & clip\_param & 0.2 \\
Value Function Clipping & use\_clipped\_value\_loss & True \\
Value Loss Coefficient & value\_loss\_coef & 1.0 \\
Adaptive KL Penalty & desired\_kl & 0.01 \\
Entropy Regularization & entropy\_coef & 0.01 \\
Generalized Advantage Estimation\\ (GAE) & gamma & 0.99 \\
& lam & 0.95 \\
Gradient Clipping & max\_grad\_norm & 1.0 \\
Learning Rate & learning\_rate & 0.001 \\
Mini-Batch & num\_mini\_batches & 4 \\
Learning Epochs & num\_learning\_epochs & 5 \\
Schedule & schedule & adaptive \\
Empirical Normalization & empirical\_normalization & True \\
\bottomrule
\end{tabular}
\end{small}
\end{center}
\vspace{-1.em}
\end{table*}

\subsubsection{Reward Function Design}
To ensure that the policy achieves the desired grasping behavior on our robot, we design a reward function that reflects the task's objectives. As grasping is complicated, it can be divided into a multi-stage reward function: first reaching the object, then closing the jaws, then picking up the object, and finally retracting the robotic arm to a specified position. 
Definitions of different part of our reward function for the specific object-grasp task are stated as follows:
\begin{itemize}
    \item \textbf{Fingers Open}
    \[
    R_{\text{fingers\_open}} = \tanh\left(\frac{\| \mathbf{p}_{\text{object}} - \mathbf{p}_{\text{ee}} \|_2}{\sigma}\right) \cdot \sum_{i=1}^2 \text{finger\_pos}[i]
    \]
    where \(\mathbf{p}_{\text{object}}\) represents the position of the object in world coordinates.  
    \(\mathbf{p}_{\text{ee}}\) represents the position of the end-effector in world coordinates.  
    \(\text{finger\_pos}\) represents the joint positions of the robot's fingers.  
    \(\sigma = 0.1\) is a scaling factor for distance normalization.  

    \item \textbf{Reaching Object}
    \begin{equation*}
    \begin{split}
        R_{\text{reaching\_object}} = & \left(1 - \tanh\left(\frac{\| \mathbf{p}_{\text{object}} - \mathbf{p}_{\text{ee}} \|_2}{\sigma}\right)\right)\\
        + & \left(1 - \tanh\left(\frac{\| \mathbf{p}_{\text{object}} - \mathbf{p}_{\text{ee}} \|_2}{\sigma / 4}\right)\right)
    \end{split}
    \end{equation*}
    where \(\mathbf{p}_{\text{object}}\) represents the position of the object in world coordinates.  
    \(\mathbf{p}_{\text{ee}}\) represents the position of the end-effector in world coordinates.  
    \(\sigma=0.2\) is a scaling factor for distance normalization.  

    \item \textbf{Object Goal Tracking}
    \begin{equation*}
    \begin{split}
        R_{\text{goal\_tracking}} = & \mathds{1}(\mathbf{p}_{\text{object},z} > z_{\text{init}} + h_{\text{min}})\\
        & \left(1 - \tanh\left(\frac{\| \mathbf{p}_{\text{goal}} - \mathbf{p}_{\text{object}} \|_2}{\sigma}\right)\right)
    \end{split}
    \end{equation*}
    where \(\mathbf{p}_{\text{goal}}\) represents the target position of the object in world coordinates.  
    \(\mathbf{p}_{\text{object}}\) represents the position of the object in world coordinates.  
    \(\mathbf{p}_{\text{object},z}\) represents the \(z\)-coordinate (height) of the object in world coordinates.
    \(z_{\text{init}}\) is the initial height of the object.  
    \(h_{\text{min}}=0.02\) is the minimal height above the initial height required to achieve the goal.  
    \(\sigma=0.3\) is a scaling factor for distance normalization.

    \item \textbf{Object Goal Tracking (Fine-Grained)}
    \begin{equation*}
    \begin{split}
        R_{\text{goal\_tracking\_fine}} = &\mathds{1}(\mathbf{p}_{\text{object},z} > z_{\text{init}} + h_{\text{min}}) \\
        & \left(1 - \tanh\left(\frac{\| \mathbf{p}_{\text{goal}} - \mathbf{p}_{\text{object}} \|_2}{\sigma_{\text{fine}}}\right)\right)
    \end{split}
    \end{equation*}
    where \(\mathbf{p}_{\text{goal}}\) represents the target position of the object in world coordinates.  
    \(\mathbf{p}_{\text{object}}\) represents the position of the object in world coordinates.  
    \(z_{\text{init}}\) is the initial height of the object.  
    \(h_{\text{min}}=0.02\) is the minimal height above the initial height required to achieve the goal.  
    \(\sigma_{\text{fine}}=0.05\) provides finer scaling for precise tracking.  

    \item \textbf{Action Rate Penalty}
    \[
    R_{\text{action\_rate}} = - \| \mathbf{a}_{t} - \mathbf{a}_{t-1} \|_2^2
    \]
    where \(\mathbf{a}_{t}\) represents the action at the current time step.  
    \(\mathbf{a}_{t-1}\) represents the action at the previous time step.  

    \item \textbf{Joint Velocity Penalty}
    \[
    R_{\text{joint\_vel}} = - \sum_{i \in \text{joint\_ids}} \dot{q}_i^2
    \]
    where \(\dot{q}_i\) represents the joint velocity of the \(i\)-th joint.  
    \(\text{joint\_ids}\) is the set of joint indices to penalize.  

    \item \textbf{Contact Forces}
    \begin{equation*}
    \begin{split}
        R_{\text{contact\_forces}} = & \mathds{1}\left(\frac{\text{finger\_dist}}{4} < \text{proj\_len} < \frac{\text{finger\_dist}}{1.25} + 0.2 \right) \\
        & \left(1.2 - \tanh\left(\frac{\text{object\_dist}}{0.2}\right)\right) \\
        & \mathds{1}\left(\|\mathbf{p}_{\text{object}} - \mathbf{p}_{\text{top}}\| < 0.1 \right) \\
        & \left(1 - \tanh\left(\frac{\|\mathbf{org\_vec} - \mathbf{y}\|}{0.3}\right)\right)
    \end{split}
    \end{equation*}
    where
    \(\mathbf{p}_{\text{top}}, \mathbf{p}_{\text{bottom}}, \mathbf{p}_{\text{left}}, \mathbf{p}_{\text{right}}\) are the positions of the robot's end-effector fingers (top, bottom, left, right) in world coordinates.  
    \(\mathbf{p}_{\text{object}}\) is the position of the object in world coordinates.  
    \(\mathbf{org\_vec} = \mathbf{p}_{\text{left}} - \mathbf{p}_{\text{right}}\) is the vector between the left and right fingers, rotated into the object's local frame.  
    \(\mathbf{y} = [0, 1, 0]\) is the unit vector along the y-axis in the object's local frame.  
    \(\text{finger\_dist} = \|\mathbf{p}_{\text{left}} - \mathbf{p}_{\text{right}}\|_2\) is the distance between the left and right fingers.  
    \(\text{proj\_len} = \frac{(\mathbf{p}_{\text{object}} - \mathbf{p}_{\text{left}}) \cdot (\mathbf{p}_{\text{right}} - \mathbf{p}_{\text{left}})}{\text{finger\_dist}}\) is the projection of the object's position onto the vector between the left and right fingers.  
    \(\text{object\_dist} = \frac{\|\mathbf{p}_{\text{object}} - \mathbf{p}_{\text{bottom}} \times (\mathbf{p}_{\text{top}} - \mathbf{p}_{\text{bottom}})\|}{0.11}\) is the perpendicular distance from the object to the line formed by the bottom and top fingers.

    \item \textbf{Close Fingers}
    \begin{equation*}
    \begin{split}
        R_{\text{close\_fingers}} = & \mathds{1}(\text{grasp\_force} > 0.5) \\
        & \left(1 - \tanh\left(\frac{\sum_{i=1}^2 \text{finger\_pos}[i]}{0.1}\right)\right) 
    \end{split}
    \end{equation*}
    where \(\text{grasp\_force}\) represents the force exerted by the fingers on the object.  
    \(\text{finger\_pos}\) represents the joint positions of the robot's fingers.  

    \item \textbf{Lift End Effector (EE)}
    \[
    R_{\text{lift\_ee}} = \frac{\text{clamp}(\mathbf{p}_{\text{ee},z} - 0.6, 0.0, 0.3)}{0.3} \cdot \mathds{1}(\text{grasp\_force} > 0.5)
    \]
    where \(\mathbf{p}_{\text{ee},z}\) represents the height of the end-effector above the ground.  

    \item \textbf{Lift Object}
    \begin{equation*}
    \begin{split}
        R_{\text{lift\_object}} = & \mathds{1}(\mathbf{p}_{\text{object},z} > z_{\text{init}} + h_{\text{min}}) \\
        & \frac{\text{clamp}(\mathbf{p}_{\text{object},z} - z_{\text{init}} - h_{\text{min}}, 0.0, 0.2)}{0.2}
    \end{split}
    \end{equation*}
    where \(\mathbf{p}_{\text{object},z}\) represents the height of the object above the ground.  
    \(z_{\text{init}}\) is the initial height of the object.  
    \(h_{\text{min}}=0.02\) is the minimal height above the initial height required to achieve the goal.  
\end{itemize}

Details of our weight design for the reward function can be found in Table~\ref{Reward function table}:
% Details of our reward function design for the specific object-grasp tasks can be found in Table~\ref{Reward function table}

\begin{table}[h]
\caption{Reward Function Design}
% \vspace{-1.em}
\label{Reward function table}
% \vskip 0.15in
\begin{center}
\begin{small}
\begin{tabular}{ll}
\toprule
\textbf{Reward Term} & \textbf{Weight}  \\
\midrule
Fingers Open & 5.0 \\
Reaching Object & 15.0 \\
Object Goal Tracking & 80.0 \\
Object Goal Tracking (Fine-Grained) & 70.0 \\
Action Rate Penalty & -1e-4 \\
Joint Velocity Penalty & -1e-4 \\
Contact Forces & 10 \\
Close Fingers & 100 \\
Lift End Effector (EE) & 20 \\
Lift Object & 100 \\
\bottomrule
\end{tabular}
\end{small}
\end{center}
\vspace{-1.em}
\end{table}

\subsection{Data Generation Details}
% Therefore, we generate large amounts of simulator trajectories that include bounding boxes, robot states, and actions at each step. 
In data generation, as shown in Fig.~\ref{fig:method}, we feed the images from the $n_{cam}$ first-person cameras of the robot to YOLO-World at each timestep to detect the $n_{obj}$ target objects to be manipulated, thus obtaining $n_{cam}*n_{obj}$ bounding boxes.
Then we normalize the bounding box, concatenate them, and obtain a vector $\vo$ of dimension $4*n_{cam}*n_{obj}$ as the visual feature input for the control policy.
We then store the visual features $\vo^{\text{vis}}_{t}$, all joint positions of the robot $\vo^{\text{rob}}_t$, and actions $\va_t$ for each step in the form of trajectories. 

To reduce visual interference between parallel environments and minimize the sim-to-real gap, we incorporate a white room during data collection. 
The room is designed to be spacious enough to accommodate the maximum working space, including robot, table, and apple.
Three RGB-D cameras\footnote{But we only use RGB information.}, calibrated with intrinsic and extrinsic parameters matching those of the real robot, capture visual information. 
During inference, the images from these cameras are processed by YOLO-World to generate object bounding boxes.

\begin{table*}[h]
\caption{Camera Parameters}
\vspace{-1.em}
\label{tab:camera_params}
% \vskip 0.15in
\begin{center}
\begin{small}
\begin{tabular}{p{3cm}p{5cm}p{4.5cm}}
\toprule
\textbf{Camera} & \textbf{Link} & \textbf{Image Shape and Type} \\
\midrule
\multirow{2}{*}{\textbf{cam\_high}} & 
\begin{tabular}[t]{@{}l@{}}
\multirow{2}{*}{fr\_link6} %\\
% Pos: $(-0.1618, 0.29248, 0.0021)$\\ 
% Rot: $(0.9997, 0, 0.0244, 0)$
\end{tabular} & 
\begin{tabular}[t]{@{}l@{}}
Shape: $[480, 640, 4]$\\ 
RGB, Range: 0-255 (torch.uint8)
\end{tabular} \\
\cmidrule{1-3}
\multirow{2}{*}{\textbf{cam\_left\_wrist}} & 
\begin{tabular}[t]{@{}l@{}}
\multirow{2}{*}{fl\_link6}%\\
% Pos: $(0.0382, -0.018, 0.10315)$\\ 
% Rot: $(0.9738, 0, 0.2152, 0.0454)$
\end{tabular} & 
\begin{tabular}[t]{@{}l@{}}
Shape: $[480, 640, 4]$\\ 
RGB, Range: 0-255 (torch.uint8)
\end{tabular} \\
\cmidrule{1-3}
\multirow{2}{*}{\textbf{cam\_right\_wrist}} & 
\begin{tabular}[t]{@{}l@{}}
\multirow{2}{*}{fr\_link6}%\\
% Pos: $(0.0482, 0, 0.11)$\\ 
% Rot: $(0.9724, 0, 0.2331, 0)$
\end{tabular} & 
\begin{tabular}[t]{@{}l@{}}
Shape: $[480, 640, 4]$\\ 
RGB, Range: 0-255 (torch.uint8)
\end{tabular} \\
\bottomrule
\end{tabular}
\end{small}
\end{center}
\vspace{-1.5em}
\end{table*}

Given the high memory demands of image data, we run 16 parallel environments on a single GPU
\footnote{RTX4090D, 24GB.}
% \footnote{RTX 4090, 24GB} 
for data generation, achieving up to 36k successful trajectories per day. 
For ACT training, these trajectories are subsequently stored in an HDF5 data structure\footnote{Video-based dataset} (see Table~\ref{tab:hdf5_structure}). In our approach, we convert images to bounding boxes using YOLO-World and store the results in a dictionary format. Each trajectory consists of multiple steps, with each step containing bbox, qpos, and action data.
\begin{table*}[h]
\caption{HDF5 Data Structure}
\label{tab:hdf5_structure}
% \vskip 0.15in
\begin{center}
\begin{small}
\begin{tabular}{lllll}
\toprule
\textbf{Group/Attribute} & \textbf{Dataset/Attribute Name} & \textbf{Description} \\
\midrule
\multirow{2}{*}{\textbf{Root Attributes}}  & sim & Boolean indicating simulation (False) \\
 & compress & Compression flag from `CollectEpsBuf` \\
\midrule
\multirow{4}{*}{\textbf{Datasets}}  & action & Array of actions taken \\
 & reward & Array of rewards received \\
& base\_action & Array of base actions \\
& base\_action\_t265 & (Optional) Base actions from T265 sensor \\
\midrule
\multirow{3}{*}{\textbf{Observations: states}} & qpos & Array of joint positions \\
 & qvel & Array of joint velocities \\
 & effort & Array of joint efforts \\
\midrule
\multirow{3}{*}{\textbf{Observations: images}}  & cam\_high & Byte-encoded images from `cam\_high` \\
& cam\_left\_wrist & Byte-encoded images from `cam\_left\_wrist` \\
 & cam\_right\_wrist & Byte-encoded images from `cam\_right\_wrist` \\
% \midrule
% \multirow{3}{*}{\textbf{Observations: depth images}} 
% & cam\_high & (Optional) Byte-encoded depth images from `cam\_high` \\
% & cam\_left\_wrist & (Optional) Byte-encoded depth images from `cam\_left\_wrist` \\
% & cam\_right\_wrist & (Optional) Byte-encoded depth images from `cam\_right\_wrist` \\
% \ycy{it seems that we donot utilize the depth images}
\bottomrule
\end{tabular}
\end{small}
\end{center}
% \vskip -0.1in
\end{table*}

\subsection{Student Policy Details}

We use RNN and ACT to train our student policy. 
The details of training parameters can be found in Table~\ref{RNN_training_params} and Table~\ref{act_training_params}.

\begin{table}[h]
\caption{RNN Training Parameters}
\label{RNN_training_params}
\begin{center}
\begin{small}
\begin{tabular}{ll}
\toprule
\textbf{Parameter} & \textbf{Value}  \\
\midrule
lr & 0.002 \\
lr\_backbone & 7e-05 \\
epochs & 50 \\
warmup\_ratio & 0.1 \\
use\_scheduler & cos \\
weight\_decay & 0.0001 \\
loss\_function & l1 \\
hidden\_dim & 512 \\
rnn\_layers & 3 \\
rnn\_hidden\_dim & 512 \\
actor\_hidden\_dim & 512 \\
policy\_class & RNN \\
gradient\_accumulation\_steps & 1 \\
random\_mask\_ratio & 0.3 \\
\bottomrule
\end{tabular}
\end{small}
\end{center}
% \vspace{-1.em}
\end{table}

\begin{table}[h]
% \vspace{-1em}
\caption{ACT Training Parameters}
% \vspace{-1em}
\label{act_training_params}
\begin{center}
\begin{small}
\begin{tabular}{ll}
\toprule
\textbf{Parameter} & \textbf{Value}  \\
\midrule
num\_epochs & 1000 \\
lr & 4e-05 \\
lr\_backbone & 7e-05 \\
weight\_decay & 0.0001 \\
kl\_weight & 10 \\
backbone & resnet18 \\
position\_embedding & sine \\
loss\_function & l1 \\
chunk\_size & 32 \\
hidden\_dim & 512 \\
dim\_feedforward & 3200 \\
enc\_layers & 4 \\
dec\_layers & 7 \\
nheads & 8 \\
dropout & 0.2 \\
train\_loader\_len & 23 \\
warmup\_ratio & 0.1 \\
scheduler & cosine \\
\bottomrule
\end{tabular}
\end{small}
\end{center}
\end{table}

\subsection{Sim2Real Technique}
\label{Sim2Real}

We employ the student policy for sim-to-real transfer, where the inputs to the policy included proprioception \(o^{rob}_{t}\) which consists of joint positions, and the visual input \(o^{vis}_{t}\), represented by bounding boxes. 
The output action is the desired joint position for the next step. 
Thus, the primary sim-to-real gaps exist in the joint position and bounding box dimensions.

For joint positions, we develop a real-to-sim code that maps the joint positions obtained from teleoperation on a real-world robot to a simulated robot. We evaluate whether the actions performed at specific target positions were consistent between the real-world and simulated environments. For example, when teleoperating to grasp an apple at a precisely measured location, both the real-world and simulated apples are set at the same coordinates.
We used the main arm of the real-world robot to observe whether the simulated robot could successfully grasp the apple.

Special handling is required for the gripper, as there are discrepancies between the URDF model of the gripper and its real-world counterpart. 
In the simulation, the gripper has two joints, while in reality, it functions as a single joint, leading to differences in value interpretation. 
To address this, we collect 10 data sets for fitting and utilize function interpolation to map the simulated gripper joint positions to those of the real-world gripper. This approach ensures that the state space and action space are aligned between the simulation and the real environment.

The sim-to-real gap for the bounding boxes corresponds to the image sim-to-real gap. 
We perform hand-eye calibration on the camera to ensure that the cameras in both the simulation and the real-world robot are approximately aligned. 
For finer adjustments, we select a specific point in front of the robot and a specific-sized object, observing the bounding boxes obtained from both the simulator and the real-world camera. 
Since we use bounding boxes, other factors, such as lighting and background variations in the Sim2Real gap are not a concern. 
Ultimately, this method of adjustment ensures that the intrinsic and extrinsic parameters of the camera are consistent between the simulation and the real-world environment.

% Other methods to ensure successful Sim2Real, like domain randomization, are discussed in the previous section (Sec.~\ref{sec_teacher}).

\section{Hardware Details}
\label{sec:hardware}
\begin{figure}[!ht]
    \centering
    \begin{minipage}[b]{0.55\textwidth}
        \centering
        \captionof{table}{Technical specifications.}
        \label{tab:tech_spec}
        \begin{tabular}{l c}
            \toprule
            Parameter & Value \\
            \midrule
            Size & 1080 $\times$ 700 $\times$ 1140 \\
            Arm weight & 4.2~kg \\ 
            Arm Payload & 3000~g (peak) \\
             & 1500~g (valid) \\
            Arm reach & 600~mm \\
            Arm repeatability & 1~mm \\
            Arm working radius &  653 mm \\ 
            Joint motion range & J1: ±154°, J2: 0°$ \sim $165° \\
            & J3: -175°$ \sim $0°, J4:  ±106°\\
            & J5: ±75° , J6: ±100° \\
            Gripper range & 0-80~mm \\
            Gripper max force & 10~NM \\
            DoF & 7 $\times$ 2 = 14 \\
            \bottomrule
        \end{tabular}
    \end{minipage}
    \hfill
    \begin{minipage}[b]{0.4\textwidth} 
        \centering
        \includegraphics[width=\textwidth]{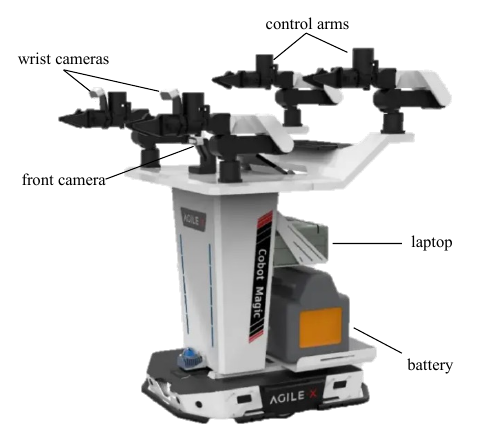}
        \caption{Hardware features.}
        \label{fig:aloha}
    \end{minipage}
\end{figure}

The Mobile ALOHA robot is equipped with a dual-arm design, providing a total of 14 degrees of freedom (DoF), with 7 DoF per arm for versatile movement. 
It owns four arms in total: two for teleoperation during data collection and two subordinate arms that perform manipulation tasks. 
Each arm has a maximum valid payload of 1500g, and the gripper has a range of up to 80mm, allowing the robot to perform a variety of grasping and manipulation tasks efficiently.
The detailed hardware information and an image of the robot can be found in Figure~\ref{fig:aloha} and Table~\ref{tab:tech_spec}.

\section{Limitations and Future Works}
\label{limitation}
One of the limitations of ManiBox is that the accuracy of the obtained bounding box depends on the capability of the preceding visual model. 
As the YOLO-World sometimes may not be able to accurately partition the bounding box, ManiBox may grasp the wrong positions due to the wrong input bounding box. 
Also, the bounding box is mainly for objects with convex shapes and may be insufficient for handling flexible objects and fluids.
In the future, a promising direction is to apply more powerful visual models like Grounding DINO and capture more semantic visual signals like segmentation via SAM in ManiBox to improve its ability.
Also, for some complicated manipulation tasks like folding up clothes, directly training an RL teacher agent in the simulator may be difficult. Thus, how to utilize human demonstration or state machines to get the expert teacher agent is also worth researching.

\section{Broader Impacts}
\label{broader impacts}
This work is about the basic theoretical algorithms of machine learning and robot control, so there is no particular broader impact to discuss.

\section{Experiments Compute Resources}
\label{experiments compute resources}
All experiments are conducted on a single NVIDIA RTX 4090D GPU (24GB) and an Intel Core i7 (9th Generation) CPU. The training of the teacher policy requires approximately 45 minutes, while the data collection process for generating 36,000 trajectories takes around 24 hours. The student policy can be trained in under 10 minutes.
For policy evaluation, generating 48 trajectories across 16 parallel environments takes approximately 30 minutes.

%%%%%%%%%%%%%%%%%%%%%%%%%%%%%%%%%%%%%%%%%%%%%%%%%%%%%%%%%%%%

\newpage
\section*{NeurIPS Paper Checklist}

\begin{enumerate}

\item {\bf Claims}
    \item[] Question: Do the main claims made in the abstract and introduction accurately reflect the paper's contributions and scope?
    \item[] Answer: \answerYes{} % Replace by \answerYes{}, \answerNo{}, or \answerNA{}.
    \item[] Justification: The main claims of the paper are clearly stated in the abstract and further elaborated at the end of the introduction. The contributions are explicitly listed and align with the theoretical and experimental results presented in the paper. The scope of the work is well defined, including its assumptions and limitations.

\item {\bf Limitations}
    \item[] Question: Does the paper discuss the limitations of the work performed by the authors?
    \item[] Answer: \answerYes{} % Replace by \answerYes{}, \answerNo{}, or \answerNA{}.
    \item[] Justification: The limitations of the work can be found in Appendix \ref{limitation}.

\item {\bf Theory assumptions and proofs}
    \item[] Question: For each theoretical result, does the paper provide the full set of assumptions and a complete (and correct) proof?
    \item[] Answer: \answerYes{} % Replace by \answerYes{}, \answerNo{}, or \answerNA{}.
    \item[] Justification: The Lemma \ref{lemma1} and Lemma \ref{lemma2} have been fully stated in the main text \ref{sec_data_collect} Simulation Data Generation. The detailed proofs are provided in Appendix \ref{app_proof1} and \ref{app_proof2}.

    \item {\bf Experimental result reproducibility}
    \item[] Question: Does the paper fully disclose all the information needed to reproduce the main experimental results of the paper to the extent that it affects the main claims and/or conclusions of the paper (regardless of whether the code and data are provided or not)?
    \item[] Answer: \answerYes{} % Replace by \answerYes{}, \answerNo{}, or \answerNA{}.
    \item[] Justification: The source code is included in the supplementary materials, accompanied by detailed instructions for setting up the environment and reproducing all key experiments. This ensures that the main results and claims of the paper can be independently verified.

\item {\bf Open access to data and code}
    \item[] Question: Does the paper provide open access to the data and code, with sufficient instructions to faithfully reproduce the main experimental results, as described in supplemental material?
    \item[] Answer: \answerYes{} % Replace by \answerYes{}, \answerNo{}, or \answerNA{}.
    \item[] Justification: Refer to source code provided in supplemental material.

\item {\bf Experimental setting/details}
    \item[] Question: Does the paper specify all the training and test details (e.g., data splits, hyperparameters, how they were chosen, type of optimizer, etc.) necessary to understand the results?
    \item[] Answer: \answerYes{} % Replace by \answerYes{}, \answerNo{}, or \answerNA{}.
    \item[] Justification: Full details are provided in the code.

\item {\bf Experiment statistical significance}
    \item[] Question: Does the paper report error bars suitably and correctly defined or other appropriate information about the statistical significance of the experiments?
    \item[] Answer: \answerYes{} % Replace by \answerYes{}, \answerNo{}, or \answerNA{}.
    \item[] Justification: We report results averaged over three random seeds and include confidence bars representing the standard error of the mean.

\item {\bf Experiments compute resources}
    \item[] Question: For each experiment, does the paper provide sufficient information on the computer resources (type of compute workers, memory, time of execution) needed to reproduce the experiments?
    \item[] Answer: \answerYes{} % Replace by \answerYes{}, \answerNo{}, or \answerNA{}.
    \item[] Justification: Details are provided in Appendix \ref{experiments compute resources}
    
\item {\bf Code of ethics}
    \item[] Question: Does the research conducted in the paper conform, in every respect, with the NeurIPS Code of Ethics \url{https://neurips.cc/public/EthicsGuidelines}?
    \item[] Answer: \answerYes{} % Replace by \answerYes{}, \answerNo{}, or \answerNA{}.
    \item[] Justification: We have reviewed the NeurIPS Code of Ethics and confirm that our work adheres to all the guidelines. Our research does not involve any known ethical risks, and we have taken care to ensure responsible use of data, models, and reproducibility.

\item {\bf Broader impacts}
    \item[] Question: Does the paper discuss both potential positive societal impacts and negative societal impacts of the work performed?
    \item[] Answer: \answerYes{} % Replace by \answerYes{}, \answerNo{}, or \answerNA{}.
    \item[] Justification: Discussed in Appendix \ref{broader impacts}
    
\item {\bf Safeguards}
    \item[] Question: Does the paper describe safeguards that have been put in place for responsible release of data or models that have a high risk for misuse (e.g., pretrained language models, image generators, or scraped datasets)?
    \item[] Answer: \answerNA{} % Replace by \answerYes{}, \answerNo{}, or \answerNA{}.
    \item[] Justification: The models and datasets used in this paper do not pose foreseeable risks of misuse or dual-use.

\item {\bf Licenses for existing assets}
    \item[] Question: Are the creators or original owners of assets (e.g., code, data, models), used in the paper, properly credited and are the license and terms of use explicitly mentioned and properly respected?
    \item[] Answer: \answerYes{} % Replace by \answerYes{}, \answerNo{}, or \answerNA{}.
    \item[] Justification: We used the IsaacLab codebase (version v1.2.0, commit e00d625), which is released under the BSD-3-Clause license. The official repository is available at \url{https://github.com/isaac-sim/IsaacLab}. We have properly credited the authors in our paper, and our use complies with the license terms.

\item {\bf New assets}
    \item[] Question: Are new assets introduced in the paper well documented and is the documentation provided alongside the assets?
    \item[] Answer: \answerYes{} % Replace by \answerYes{}, \answerNo{}, or \answerNA{}.
    \item[] Justification: The new assets are included in the code provided in supplementary materials.

\item {\bf Crowdsourcing and research with human subjects}
    \item[] Question: For crowdsourcing experiments and research with human subjects, does the paper include the full text of instructions given to participants and screenshots, if applicable, as well as details about compensation (if any)? 
    \item[] Answer: \answerNA{} % Replace by \answerYes{}, \answerNo{}, or \answerNA{}.
    \item[] Justification: The paper does not involve crowdsourcing nor research with human subjects.

\item {\bf Institutional review board (IRB) approvals or equivalent for research with human subjects}
    \item[] Question: Does the paper describe potential risks incurred by study participants, whether such risks were disclosed to the subjects, and whether Institutional Review Board (IRB) approvals (or an equivalent approval/review based on the requirements of your country or institution) were obtained?
    \item[] Answer: \answerNA{} % Replace by \answerYes{}, \answerNo{}, or \answerNA{}.
    \item[] Justification: This work does not involve human subjects or crowdsourced data.

\item {\bf Declaration of LLM usage}
    \item[] Question: Does the paper describe the usage of LLMs if it is an important, original, or non-standard component of the core methods in this research? Note that if the LLM is used only for writing, editing, or formatting purposes and does not impact the core methodology, scientific rigorousness, or originality of the research, declaration is not required.
    %this research? 
    \item[] Answer: \answerNA{} % Replace by \answerYes{}, \answerNo{}, or \answerNA{}.
    \item[] Justification: The core method development in this research does not
involve LLMs as any important, original, or non-standard components.

\end{enumerate}

\end{document}